\documentclass[mathfont=cm,accepted]{uai2021} 

\usepackage[american]{babel}

\usepackage{natbib}
\usepackage{mathtools}
\usepackage{amsfonts}       
\usepackage{amsmath}
\usepackage{amssymb}
\usepackage{nicefrac}
\usepackage[separate-uncertainty = true]{siunitx}
\usepackage{booktabs}
\usepackage{makecell}
\usepackage{multirow}
\usepackage{subfigure}
\usepackage{mathtools}
\usepackage{dblfloatfix}
\usepackage{csquotes}
\usepackage[capitalise]{cleveref}
\usepackage{etoolbox}
\robustify\bfseries
\robustify\itshape

\title{\textsc{GP-ConvCNP}: Better Generalization for Convolutional Conditional Neural Processes on Time Series Data}

\author[1]{\href{mailto:Jens Petersen <jens.petersen@dkfz.de>?Subject=Your UAI 2021 paper}{Jens~Petersen}{}}
\author[1]{Gregor~Köhler}
\author[1]{David~Zimmerer}
\author[2]{Fabian~Isensee}
\author[3]{Paul~F.~Jäger}
\author[1]{Klaus~H.~Maier-Hein}
\affil[1]{%
    Division of Medical Image Computing\\
    German Cancer Research Center\\
    Heidelberg, Germany
}
\affil[2]{%
    HIP Applied Computer Vision Lab\\
    Division of Medical Image Computing\\
    German Cancer Research Center
}
\affil[3]{%
    Interactive Machine Learning Group\\
    German Cancer Research Center
}

\begin{document}
\maketitle

\begin{abstract}
Neural Processes (NPs) are a family of conditional generative models that are able to model a distribution over functions, in a way that allows them to perform predictions at test time conditioned on a number of context points. A recent addition to this family, \emph{Convolutional Conditional Neural Processes} (\textsc{ConvCNP}), have shown remarkable improvement in performance over prior art, but we find that they sometimes struggle to generalize when applied to time series data. In particular, they are not robust to distribution shifts and fail to extrapolate observed patterns into the future. By incorporating a Gaussian Process into the model, we are able to remedy this and at the same time improve performance within distribution. As an added benefit, the Gaussian Process reintroduces the possibility to sample from the model, a key feature of other members in the NP family.
\end{abstract}

\section{Introduction}
\label{sec:introduction}

Neural Processes \citep{garnelo_conditional_2018,garnelo_neural_2018} have been proposed as a way to leverage the expressiveness of neural networks to learn a distribution over functions (often referred to as a \emph{stochastic process}), so that they can condition their predictions on observations given at test time, a so-called \emph{context}. But what does it mean to successfully learn such a distribution? We believe that it should be characterized by the following: 1) accurate predictions, meaning predictions should be as close as possible to the true underlying function, 2) good reconstruction of the given observations, 3) generalization, because we assume that there will be some underlying generative process from which the distribution originates and which is valid beyond the finite data we observe. The latter is especially important when only few context observations are given that could be explained by several different functions. Follow-up work to Neural Processes has mostly emphasized the first two aspects, the most prominent of which are \emph{Attentive Neural Processes} (ANP) \citep{kim_attentive_2019} and \emph{Convolutional Conditional Neural Process} (\textsc{ConvCNP}) \citep{gordon_convolutional_2019}, each improving upon its predecessor in terms of both prediction accuracy and reconstruction ability.

We propose a model that addresses all of the above, with a particular focus on the ability to generalize. By combining \textsc{ConvCNP} with a Gaussian Process, we achieve a significant improvement in generalization: the model, which we call \textsc{GP-ConvCNP}, can better extrapolate far from the provided context observations---meaning into future given past and present observations---and is more robust to a distribution shift at test time. It further reintroduces the ability to sample from the model, something that \textsc{ConvCNP} is incapable of, showing a better sample distribution than both NP and ANP. Finally, we find that our proposed model often yields a significant improvement in predictive performance on in-distribution data as well. We focus our evaluation on time series data, where we see the greatest potential for applications of our model. In this context, we consider several synthetic datasets as well as real time series, specifically weather data and predator-prey population dynamics. We provide a complete implementation\footnote{\url{https://github.com/MIC-DKFZ/gpconvcnp}}, including data for convenience, to reproduce all experiments in this work.

\begin{figure*}[t]
    \centering
    \includegraphics[width=\textwidth]{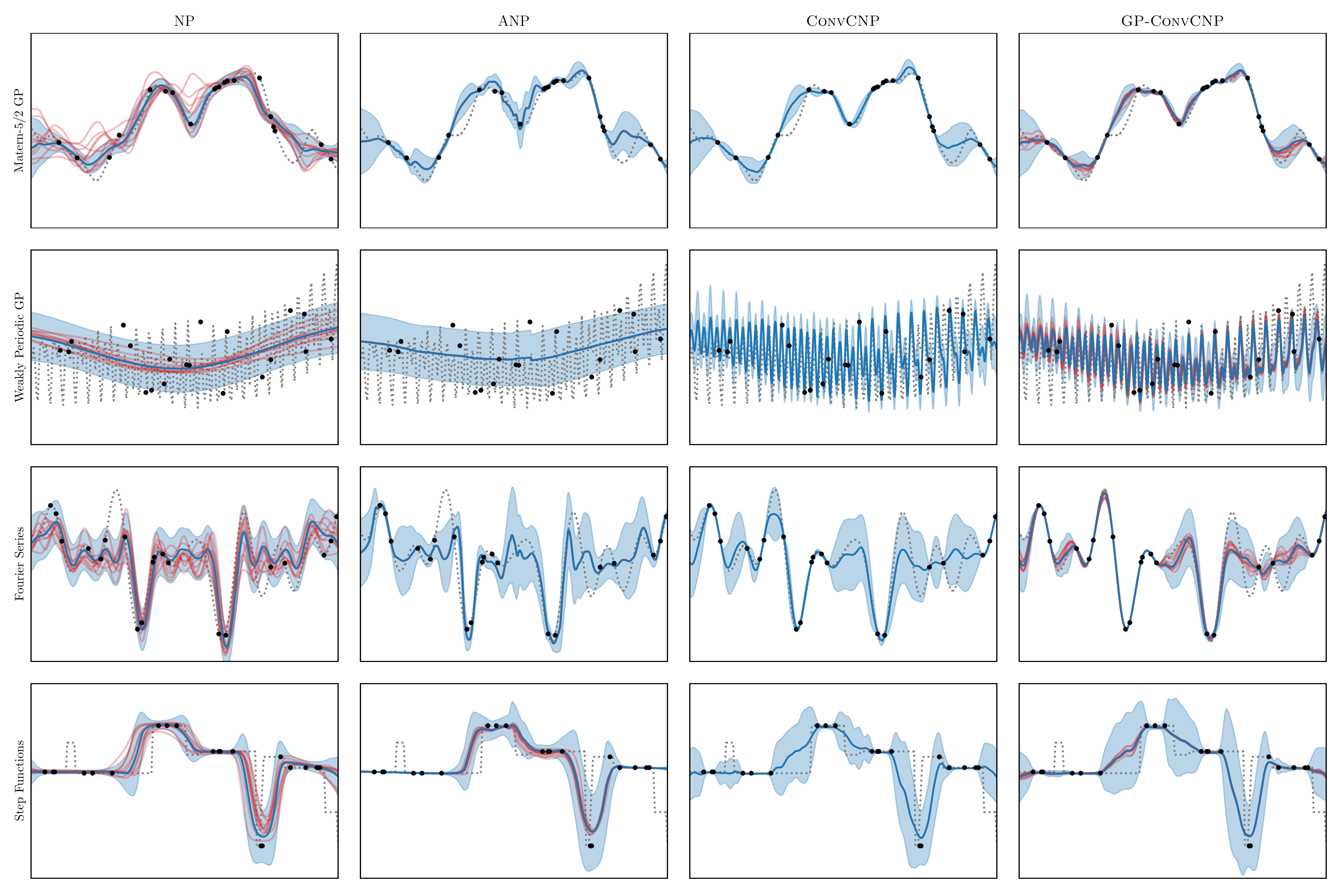}
    \caption{Our work proposes \textsc{GP-ConvCNP}, an extension of \textsc{ConvCNP} that reintroduces sampling and improves generalization on time series data. Shown here are examples for the different synthetic time series and methods evaluated in this work (mean prediction in blue, samples in red). While the mean predictions from \textsc{ConvCNP} and \textsc{GP-ConvCNP} look similar---and significantly better than those from Neural Processes (NP) and Attentive Neural Processes (ANP)---only \textsc{GP-ConvCNP} combines high quality predictions (a feature of \textsc{ConvCNP}) with the ability to sample (a feature of NP and ANP). While synthetic data measures in-distribution performance, we evaluate generalization capabilities on real data.}
    \label{fig:synthetic}
\end{figure*}

\section{Problem Statement \&\ Methods}
\label{sec:methods}

In the framework of Neural Processes \citep{garnelo_conditional_2018,garnelo_neural_2018} we assume that we are given a set of $N$ observations $C=\{(x_c,y_c)\}_{c=1}^N\eqqcolon(\mathbf{x_c},\mathbf{y_c})$, often called the \emph{context}, where $x_c\in X$ are samples from the input space $X$ and $y_c\in Y$ are samples from the output space $Y$ (commonly $X=\mathbb{R}^{d_X}$ and $Y=\mathbb{R}^{d_Y}$, in this work we restrict ourselves to $X=\mathbb{R}$, because time is scalar). It is assumed that these observations were generated by some function $f:X\rightarrow Y$, i.e. $y_c=f(x_c)$, and our goal is to infer $f$ from $C$ so that we may evaluate it at arbitrary new input locations $x_t$. In reality, this will most likely mean we have collected a number of measurements over time and are interested in an $f$ that lets us interpolate and extrapolate those measurements. Note that when we speak of predictive performance, we refer to both of those cases and not in a temporal sense. The problem is ill-posed without placing further assumptions on $f$, which is why we typically restrict it to some family $\mathcal{F}$: polynomials of some order, a combination of oscillating functions with different frequencies, etc.. However, in many cases it is undesired or even impossible to manually specify $\mathcal{F}$, so Neural Processes propose to use neural networks to learn an approximate representation of $\mathcal{F}$ by observing many examples $f\in\mathcal{F}$. The latter are typically represented as a context set $C$ (the measurements we have) and a \emph{target} set $T=\{(x_t,y_t)\}_{t=1}^M\eqqcolon(\mathbf{x_t},\mathbf{y_t})$ (the measurements we're interested in). By learning to reconstruct the examples $f$ from a limited number of context points a model should implicitly form a representation of $\mathcal{F}$, which leads to the following learning objective:

\begin{align}
    &\max_\theta\sum_{f\in\mathcal{F}}\log p_\theta(\mathbf{y_t}|\mathbf{x_t},\mathbf{x_c},\mathbf{y_c})\\
    =&\max_\theta\sum_{f\in\mathcal{F}}\sum_t\log\mathcal{N}(y_t; g^\mu_\theta(Z, x_t), g^\sigma_\theta(Z, x_t))
\label{eq:objective}
\end{align}

This objective is common to all approaches we evaluate in our work, and the second line formalizes the fact that we choose to always model the output as a diagonal Gaussian, parametrized by mean and variance functions $g^\mu_\theta,g^\sigma_\theta$ that seek to maximize the log-likelihood of the targets $\mathbf{y_t}$. The output variance can also be fixed, but \cite{le_empirical_2018} show that a learned output variance is preferable. $Z$ is a representation of the context $(\mathbf{x_c}, \mathbf{y_c})$, i.e. there is a mapping $E:X,Y\rightarrow Z$. The implementation of $E$ is where the members of the Neural Process family differ most, and we visualize them in \cref{fig:appendix:methods}.

\subsection{(Attentive) Neural Processes}
\label{sub:neuralprocess}

The original Neural Processes \citep{garnelo_neural_2018} implement $E$ as a neural network that \emph{encodes} individual context observations $(\mathbf{x_c},\mathbf{y_c})$ into a finite-dimensional space. These representations are then averaged to form the global representation $Z$. Similar to \cref{eq:objective}, $Z$ parametrizes a Gaussian distribution, which enables NP to sample from this latent space and produce diverse predictions; we do not consider the deterministic NP variant \citep{garnelo_conditional_2018} in this work. NP are trained by maximizing a lower bound on \cref{eq:objective}, similar to variational autoencoders. In our NP implementation $E$ and $(g^\mu_\theta,g^\sigma_\theta)$ are symmetric 6-layer MLP, with a representation size of 128. Attentive Neural Processes \citep{kim_attentive_2019} are motivated by the observation that NP poorly reconstruct the provided context, i.e. the predictions seem to miss the context points, as seen for example in \cref{fig:synthetic}. To mitigate this effect, ANP augment NP with an additional deterministic encoder-decoder path. Instead of averaging the individual representations, a learned attention mechanism combines them, conditioned on a target point $x_t$. So while NP need to compress representations to a single point in $Z$, ANP don't have this bottleneck, which likely contributes to their improved performance. In our ANP implementation, the deterministic path mirrors the variational path, with both the representation dimension and the embedding dimension of the attention mechanism being 128. \cite{le_empirical_2018} evaluated several hyperparameter configurations for NP and ANP and our implementation matches their best performing one.

\begin{table*}[b]
    \caption{Results for synthetically created data. Test data was generated with the same parameters as the training data, so we're looking at \emph{in-distribution} performance. $\uparrow$/$\downarrow$ indicate that higher/lower is better. Errors represent 1 standard deviation over 5 runs with different seeds (standard error of the mean for GPs, because seed influence is negligible), where each run was evaluated with \SI{102400} (\SI{30720} for $W_2$) samples. Bold indicates that the method(s) are significantly better than all non-bold methods, i.e. when the difference is larger than the root sum of squares of the standard deviations. Overall, \textsc{GP-ConvCNP} outperforms the competing approaches, especially in terms of predictive log-likelihood and sample diversity (compared to an oracle) where applicable. In terms of reconstruction error, our method outperforms prior art on three datasets, but is on par with \textsc{ConvCNP} on two of those. Interestingly, the EQ-GP, which is what our model uses as an initial estimate, performs rather poorly in all but the first example. In the first example, where the EQ-GP is already a decent estimate, our approach leverages that information and matches the oracle GP in predictive performance! The reconstruction error and $W_2$ of the oracle are zero, so we don't show them here. The dependence of model performance on the number of context points is visualized in \cref{fig:appendix:synthetic_numcontext} for the two GP examples.
    }
    \label{tab:synthetic}
    \centering
    \sisetup{detect-weight=true,detect-inline-weight=math}
    \begin{tabular}{llS[table-format=2.3(3)]S[table-format=1.3(3)]S[table-format=1.3(3)]S[table-format=1.3(3)]}
    \toprule
    & & \multicolumn{1}{c}{Matern-5/2 GP} & {Weakly Per. GP} & {Fourier Series} & {Step Functions} \\
    \midrule
    \multirow{6}{*}{Predictive LL$\uparrow$}
    & GP (EQ) & 1.031(75) & -8.034(2260) & -0.241(752) & \num{-2e17} \\
    & GP (Oracle) & 1.933(95) & 1.876(26) & & \\
    \cline{2-6}
    \rule{0pt}{1em}
    & NP & -0.496(27) & -1.161(7) & -1.743(20) & -3.287(491) \\
    & ANP & 0.723(46) & -1.047(8) & -0.976(28) & -65.141(60979) \\
    & \textsc{ConvCNP} & 1.710(38) & -0.153(33) & 0.372(65) & \bfseries -0.522(163) \\
    & \textsc{GP-ConvCNP} & \bfseries 1.930(31) & \bfseries -0.090(21) & \bfseries 1.632(79) & \bfseries -0.532(44) \\
    \midrule
    \multirow{5}{*}{Recon. Error$\downarrow$}
    & GP (EQ) & 0.001(1) & 0.028(1) & 0.004(1) & 0.097(1) \\
    \cline{2-6}
    \rule{0pt}{1em}
    & NP & 0.027(1) & 0.500(3) & 0.845(74) & 0.292(10) \\
    & ANP & \bfseries 0.008(2) & 0.491(4) & 0.181(18) & 0.284(13) \\
    & \textsc{ConvCNP} & 0.025(20) & 0.109(77) & \bfseries 0.042(27) & \bfseries 0.121(17) \\
    & \textsc{GP-ConvCNP} & 0.013(2) & \bfseries 0.061(7) & \bfseries 0.040(23) & \bfseries 0.116(17) \\
    \midrule
    \multirow{5}{*}{$W_2\downarrow$}
    & GP (EQ) & 4.294(7) & 4.521(3) & & \\
    \cline{2-6}
    \rule{0pt}{1em}
    & NP & 1.836(21) & 2.745(4) & & \\
    & ANP & 1.369(48) & 2.708(2) & & \\
    & \textsc{ConvCNP} &  &  & & \\
    & \textsc{GP-ConvCNP} & \bfseries 0.987(86) & \bfseries 1.800(45) & & \\
    \bottomrule
    \end{tabular}
\end{table*}

\subsection{From \textsc{ConvCNP} to \textsc{GP-ConvCNP}}
\label{sub:convcnp}

With the goal of enabling translation equivariance (i.e. independence of the value range of $\mathbf{x_c}$ and $\mathbf{x_t}$) in Neural Processes, the authors of Convolutional Conditional Neural Processes (\textsc{ConvCNP}) \citep{gordon_convolutional_2019} approach their work from the perspective of \emph{learning on sets} \citep{zaheer_deep_2017}. While NP and ANP map the context set into a finite-dimensional representation, \textsc{ConvCNP} map it into an infinite-dimensional function space. The authors show that in this scenario translation equivariance (as well as permutation invariance) can only be achieved if the mapping $E$ can be represented in the form

\begin{align}
    E(\mathbf{x_c},\mathbf{y_c})&=\rho(E^\prime(\mathbf{x_c},\mathbf{y_c}))\\
    E^\prime(\mathbf{x_c},\mathbf{y_c})&=\sum_c\phi(y_c)\psi(\cdot-x_c)
\end{align}

where $\phi:Y\rightarrow\mathbb{R}^2$ and $\psi:X\rightarrow\mathbb{R}$, so that $E^\prime$ defines a function and $\rho$ operates in function space and must be translation equivariant. The similar naming of $E,E^\prime$ is deliberate, because herein lies a key difference to NP (and also ANP): NP learn a powerful mapping (i.e. neural network) from the context to a representation and then another one from this representation to the output space, whereas \textsc{ConvCNP} employs a very simple mapping to another representation (to function space, because $\phi$ and $\psi$ are defined with kernels, see below). A powerful approximator is then learned that operates \emph{within} this representation space, as $\rho$ is a CNN operating on a discretization of $E^\prime$. The mapping back to output space is again a simple one, usually also $\psi$ combined with a linear map. In this sense, both $E$ and $E^\prime$ can be thought of as representations when we make the connection to NP. See also \cref{fig:appendix:methods} for a visualization of these differences. In \cite{gordon_convolutional_2019}, $\psi$ is chosen to be a simple Gaussian kernel, and $\phi$ such that the resulting $E^\prime$ has two components:

\begin{equation}
    E'(\mathbf{x_c},\mathbf{y_c})=\left(\sum_c k(\cdot, x_c)\;,\;\sum_c\frac{y_c k(\cdot, x_c)}{\sum_{c'} k(\cdot, x_{c'})}\right)
    \label{eq:nadaraya}
\end{equation}

which is the combination of a kernel density estimator and a Nadaraya-Watson estimator. 
This estimate is discretized on a suitable grid and a CNN $\rho$ is applied, the result of which is again turned into a continuous function by convolving with the (Gaussian) kernel $\psi$. We use the official implementation\footnote{\url{https://github.com/cambridge-mlg/convcnp}} in our experiments. Note that $k$ in \cref{eq:nadaraya} is the same as $\psi$ in the implementation.

In this work, we propose \textsc{GP-ConvCNP}, a model that replaces the deterministic kernel density estimate $E'$ in \textsc{ConvCNP} with a Gaussian Process posterior \citep{rasmussen_gaussian_2006}. Gaussian Processes (GP) are a popular choice for time series analysis \citep{roberts_gaussian_2013}, but typically require a lot of prior knowledge about a problem to choose an appropriate kernel. We will find that this is not the case for \textsc{GP-ConvCNP}, which is even able to learn periodicity when the chosen kernel is not periodic. 

The posterior in a GP is a normal distribution with a mean function $m(\mathbf{x_t})$ conditioned on the context and a covariance function $K(\mathbf{x_t})$ specified by some kernel $k$:

\begin{align}
    m(\mathbf{x_t})&=k_{tc}^T\left(k_{cc}+\sigma^2\mathbb{I}\right)^{-1}\mathbf{y_c}\label{eq:gp_mean}\\
    K(\mathbf{x_t})&=k_{tt}+\sigma^2-k_{tc}^T\left(k_{cc}+\sigma^2\mathbb{I}\right)^{-1}k_{tc}\label{eq:gp_cov}
\end{align}

where $k_{tc}=k(\mathbf{x_t},\mathbf{x_c})$ etc. and $\sigma^2$ is a noise parameter that essentially determines how close the prediction will be to the context points. We make this parameter learnable. Note that \cref{eq:gp_mean} is very similar to \cref{eq:nadaraya}: it corresponds to the second component of the Nadaraya-Watson estimator with only a changed denominator.

The first obvious benefit of this model is that we can sample from the GP posterior distribution and thus also from our model, recovering one very compelling property of NP that \textsc{ConvCNP} lacks. Another advantage we see is that by working with a distribution instead of a deterministic estimate as input to the CNN, the data distribution is implicitly smoothed. It has been established that such smoothing reduces overfitting and improves generalization, e.g. by adding noise to inputs \citep[p.347]{bishop_1995} or more generally doing data augmentation \citep{volpi_2018}. Working with a distribution instead of a deterministic estimate, we need to perform Monte-Carlo integration to get a prediction from our model. During training, however, we only use a single sample, as is commonly done e.g. in variational autoencoders when training with mini-batch stochastic gradient descent. To facilitate comparison, the kernel we use in our GP is the same as in \textsc{ConvCNP}, i.e. a Gaussian kernel with a learnable length scale.

Note that our model retains all desirable characteristics of the competing approaches, in particular permutation invariance with respect to the inputs (present in all prior art) and translation equivariance (present in \textsc{ConvCNP})\footnote{As in \textsc{ConvCNP}, this obviously requires a stationary kernel.}. For details on the various optimization parameters etc. we refer to the provided implementation.

\begin{figure*}[t]
    \centering
    \includegraphics[width=\textwidth]{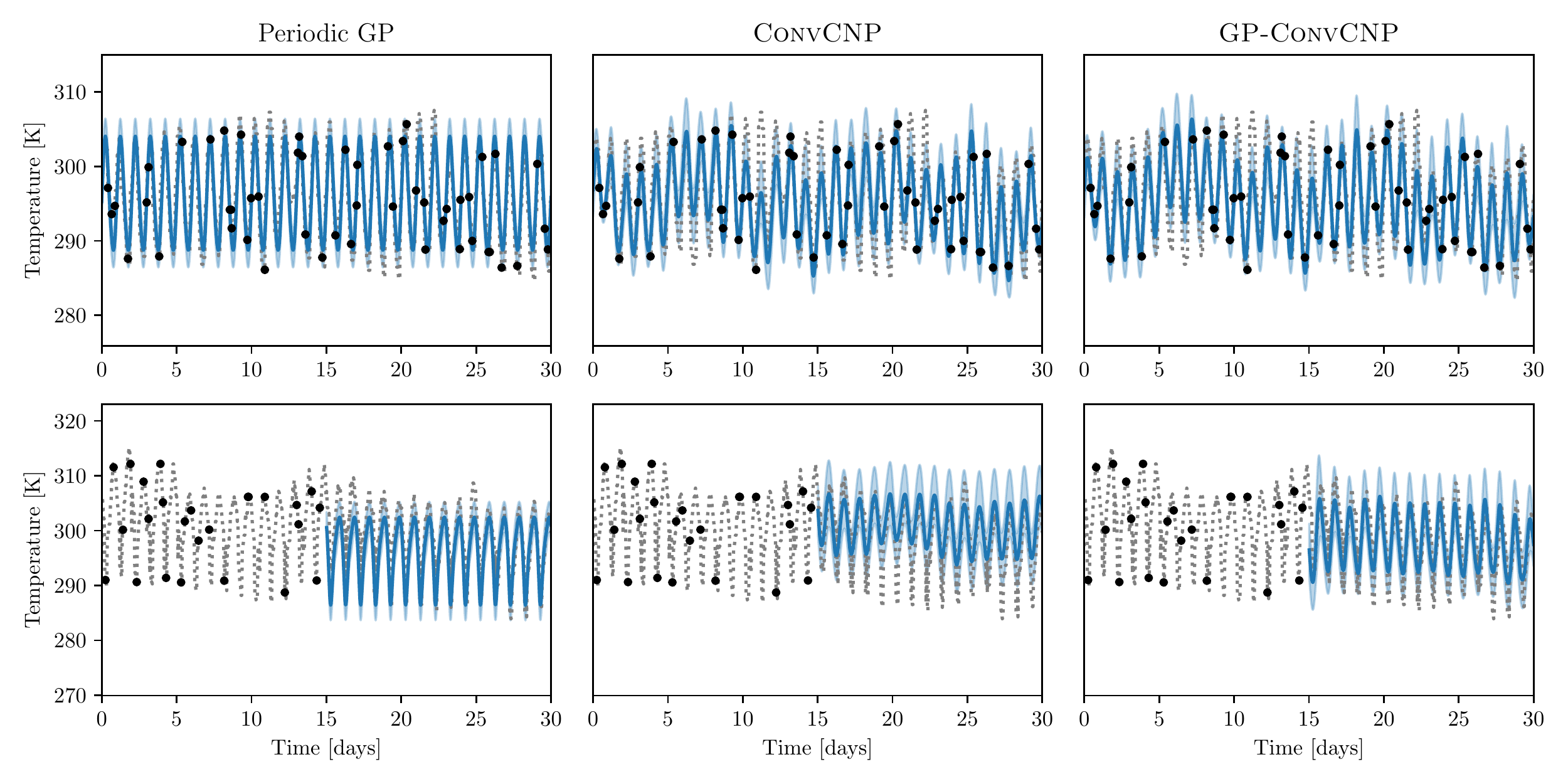}
    \caption{Examples from the temperature time series test set. For the interpolation task (top) we provide context points from the full sequence, for the extrapolation task (bottom) we provide context points in the first half of the sequence and evaluate the second. Both \textsc{ConvCNP} and \textsc{GP-ConvCNP} capture the periodicity of day/night changes in temperature well and are able to extrapolate it. We find that \textsc{GP-ConvCNP} often better matches the amplitude of the true signal, especially for the extrapolation task, which could explain its superior performance in \cref{tab:real}. Because the temperature signal is periodic, we also show a periodic GP with a commonly used Exponential Sine-Squared kernel. NP and ANP are only shown in the appendix in \cref{fig:appendix:temperature_np}, as they were unable to fit the data (similar to the weakly periodic GP data in \cref{fig:synthetic}).}
    \label{fig:temperature}
\end{figure*}

\begin{table*}[t]
    \caption{Results on the real world datasets. Again, $\uparrow$/$\downarrow$ indicate that higher/lower is better and errors represent 1 standard deviation over 5 runs with different seeds (standard error of the mean for GP, because seed influence is negligible). (left) For the temperature interpolation task, context points are randomly sampled from the test interval, for the temperature extrapolation task we provide context points in the first half of the interval and measure performance on the second half (as seen in \cref{fig:temperature}). For comparison, we also show a periodic GP with an Exponential Sine-Squared kernel. 
    (right) For the population dynamics, models are trained on simulated data, so the real world data (also shown in \cref{fig:population}) is likely out-of-distribution, as evidenced by the stark drop in performance. There is no obvious choice of kernel if one wanted to apply a GP to this problem. 
    }
    \label{tab:real}
    \centering
    \sisetup{detect-weight=true,detect-inline-weight=math}
    \begin{tabular}{llS[table-format=2.3(3)]S[table-format=1.3(3)]S[table-format=2.3(3)]S[table-format=1.3(3)]}
    \toprule
    & & \multicolumn{2}{c}{Temperature Time Series} & \multicolumn{2}{c}{Population Dynamics} \\
    \midrule
    & & \multicolumn{1}{c}{interpolation} & \multicolumn{1}{c}{extrapolation} & \multicolumn{1}{c}{simulated} & \multicolumn{1}{c}{real} \\

    \midrule \multirow{5}{*}{Predictive LL$\uparrow$}
    & GP (per.) & -2.075(237) & -46.611(2557) & & \\
    \cline{2-6}
    \rule{0pt}{1em}
    & NP & -0.855(3) & -1.267(11) & 0.527(51) & -33.070(7636) \\
    & ANP & -0.733(8) & -1.938(381) & 1.027(33) & -29.714(9210) \\
    & \textsc{ConvCNP} & \bfseries -0.522(8) & -1.261(62) & \bfseries 1.374(17) & -23.540(12441) \\
    & \textsc{GP-ConvCNP} & \bfseries -0.515(19) & \bfseries -1.190(16) & 1.337(29) & \bfseries -5.382(2625) \\
    
    \midrule  \multirow{5}{*}{Recon. Error$\downarrow$}
    & GP (per.) & 0.274(1) & & & \\
    \cline{2-6}
    \rule{0pt}{1em}
    & NP & 0.238(2) & & 0.018(1) & 1.053(15) \\
    & ANP & 0.198(7) & & 0.008(4) & 0.772(20) \\
    & \textsc{ConvCNP} & \bfseries 0.106(2) & & \bfseries 0.002(1) & \bfseries 0.374(19) \\
    & \textsc{GP-ConvCNP} & \bfseries 0.123(18) & & 0.004(1) & \bfseries 0.411(26) \\
    \bottomrule
    \end{tabular}
\end{table*}

\section{Experiments}
\label{sec:experiments}

We design our experiments with the purpose of evaluating how well members of the Neural Process family, including the one we propose, are suited for the task of learning distributions over functions, i.e. stochastic processes, specifically for time series data. Like the works we compare ourselves with, we evaluate both predictive performance (\emph{How good is our prediction between context points?}) via the predictive log-likelihood and the reconstruction performance (\emph{How good is our prediction at the context points?}) via the root-mean-square error (RSME), because predictions directly at the context points are usually extremely narrow Gaussians, leading to unstable likelihoods.

As outlined in the introduction, one defining aspect of successfully learning a distribution over functions is a model's ability to generalize. This can mean several things, for example independence with respect to the input value range, called translation equivariance. This is a key feature of \textsc{ConvCNP} (as long as a stationary kernel is used for interpolation), and we retain this property in \textsc{GP-ConvCNP}. We evaluate two further attributes of generalization, both on real world data: one is the ability to extrapolate the context information, i.e. to produce good predictions well into the future by inferring an underlying pattern; the other is the ability to deal with a distribution shift at test time, in our case a shift from simulated to real world data.

On top of the above, we are also interested in how well the distribution of samples from a model matches the ideal distribution. In general, the latter is not accessible, but for some synthetic examples we describe below, specifically those from a Gaussian Process, we do have access, simply by using the generating GP as an oracle. We can then compare this reference---a Gaussian distribution---with the distribution of samples from our model. Note that one sample is a prediction at all target points at once, as seen for example in \cref{fig:synthetic}. The majority of approaches that estimate differences between distributions fall into the categories of either \emph{$f$-divergences} or \emph{Integral Probability Measures} (for an overview see for example \cite{sriperumbudur_integral_2009}). The former require evaluations of likelihoods for both distributions, while we only have individual samples from our model. We opt for a parameter-free representative of the IPM category, the Wasserstein distance $W_2$. We elaborate further on the definition and motivation in \cref{sub:appendix:wasserstein}.

We initially test our method on diverse synthetic time series. The first two have also been used in \cite{gordon_convolutional_2019}, and they allow us to evaluate the sample diversity, as outlined above: (1) Samples from a Gaussian Process with a Matern-5/2 kernel. (2) Samples from a Gaussian Process with a weakly periodic kernel. (3) Fourier series with a variable number of components, each of which has random bias, amplitude and phase. (4) Step functions, which were specifically chosen to challenge our model, as the kernel we employ introduces smoothness assumptions that are ill-suited for this problem. All of these are described in greater detail in \cref{sec:appendix:dataeval} as well as the provided implementation. The size $N$ of the context set is drawn uniformly from $[3,100)$ and the size $M$ of the target set from $[N,100)$ following \cite{le_empirical_2018}. We further join the context set into the target set as done in \cite{garnelo_conditional_2018,garnelo_neural_2018}. Examples can be seen in \cref{fig:synthetic}.

The first real world dataset we look at are weather recordings for several different US, Canadian and Israeli cities. In particular we focus on temperature measurements in hourly intervals that have been collected over the course of 5 years (see \cref{sub:appendix:temperature}). Temperatures in each city are normalized by their respective means and standard deviations. We randomly sample sequences of $\sim$1 month as instances and evaluate two tasks, taking US and Canadian cities as the training set and Israeli cities as the test set:

\begin{enumerate}
    \item Interpolation, where we draw context points and target points randomly from the entire sequence (i.e. the same as in the synthetic examples).
    \item Extrapolation, where context points are drawn from the first half of the sequence and performance is evaluated on the second half (as shown in \cref{fig:temperature}). We can reasonably be sure that temperature changes between day and night occur in the future with the same frequency, so extrapolating this pattern is a good test of a model's ability to generalize.
\end{enumerate}

The second real world dataset are measurements of a predator-prey population of lynx and hare.
Such population dynamics are often approximated by Lotka-Volterra equations \citep{leigh_1968}, so we train models on simulated population dynamics and test on both the simulated and real world data. \cite{gordon_convolutional_2019} used this dataset as well, but only to qualitatively show that \textsc{ConvCNP} can be applied to it. The analysis will allow us to quantify how robust the models are to a shift in distribution at test time, as the simulation parameters are almost certainly not an ideal fit for the real world data. For details on the simulation process we refer to \cref{sub:appendix:population}.

Finally, even though the focus of our work is on time series data, we include some image experiments, mainly for the purpose of a more nuanced direct comparison with \textsc{ConvCNP}. In particular, we compare the models on MNIST \citep{lecun_gradient-based_1998}, CIFAR10 \citep{krizhevsky_learning-multiple_2009} and CelebA \citep{liufaceattributes_2015}. For the latter two, we work on resampled versions at $32^2$ resolution. More details are given in \cref{sec:appendix:imageexperiments}.

\begin{figure*}[t]
    \centering
    \includegraphics[width=\textwidth]{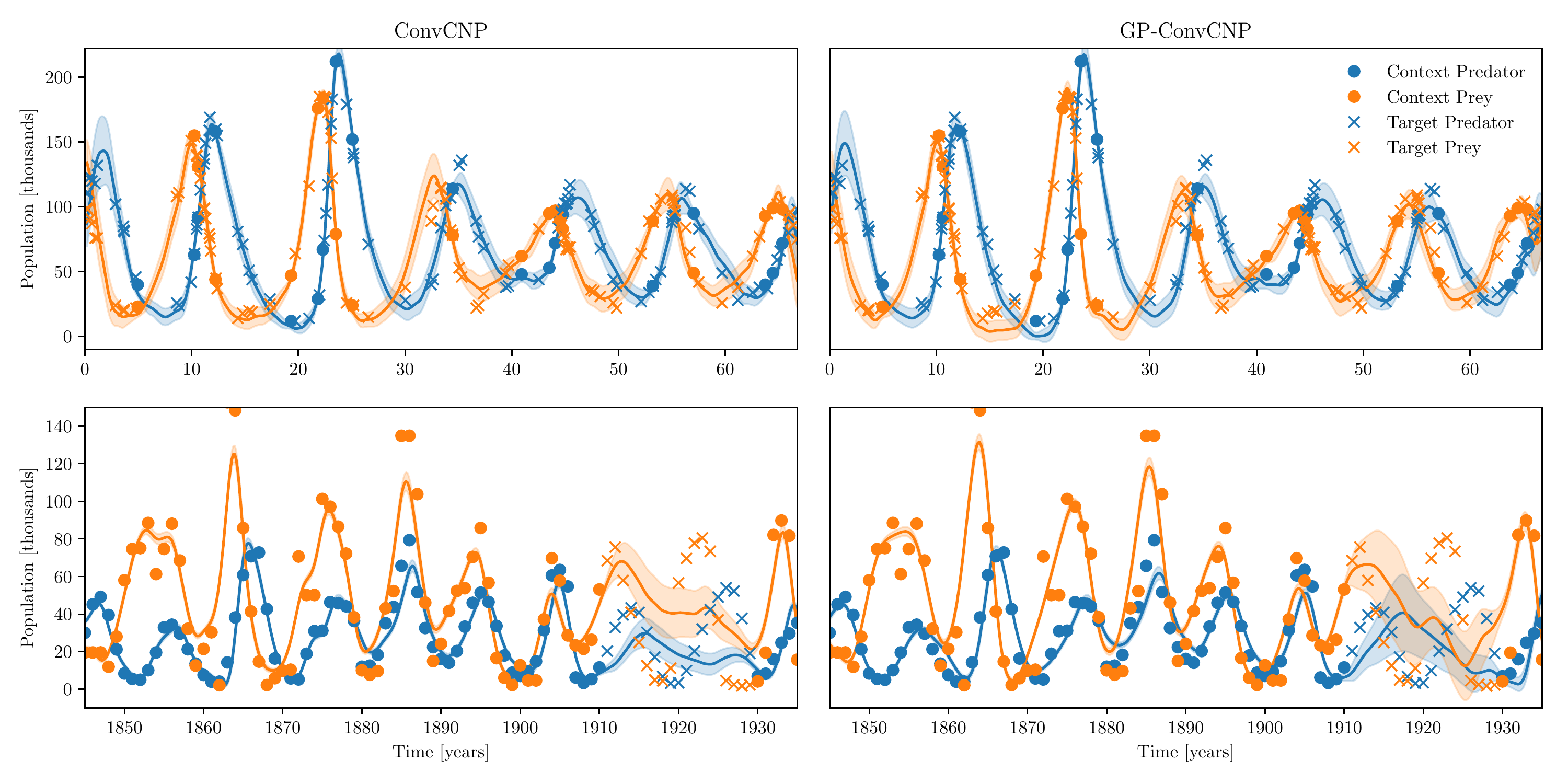}
    \caption{Example of \textsc{ConvCNP} and \textsc{GP-ConvCNP} applied to the simulated Lotka-Volterra population dynamics (top) and to the real Hudson Bay Company lynx-hare dataset (bottom). Both perform well on the simulated (i.e. in-distribution) data and seem to struggle fitting the test interval on the real world data. Not however how the predicted uncertainty is larger for \textsc{GP-ConvCNP}. We display the best out of 5 models in each case, and for \textsc{ConvCNP} the performance is much more volatile, as seen in \cref{tab:real}. NP and ANP perform poorly on the real world data, the corresponding figure is \cref{fig:appendix:population_np}.}
    \label{fig:population}
\end{figure*}

\section{Results}
\label{sec:results}

\cref{tab:synthetic} shows results for the various synthetic time series. In this experiment the models are trained and tested on random samples generated in the same way, so these results measure \emph{in-distribution} performance. We find that \textsc{GP-ConvCNP} is the overall best performing method, significantly so in terms of predictive performance for 3 out of the 4 time series and performing on par with \textsc{ConvCNP} on the other. Reconstruction performance is on par with \textsc{ConvCNP} in 3 out of 4 instances and significantly better in one. For reference, we also show results for a Gaussian Process with EQ kernel (what our model uses) and the oracle where available. Evidently, the initial GP estimate in our model doesn't have to be very good, but when it is, like in the Matern-5/2 case, our approach leverages this and even matches the oracle in performance. For examples originating from a Gaussian Process, we can evaluate the sample diversity with respect to the oracle GP, finding that \textsc{GP-ConvCNP} significantly outperforms the other methods in this regard. It is important to note, however, that this measure does not fully isolate the sample diversity. A low reconstruction error, for example, will also improve the $W_2$, which is likely the reason that ANP still performs better than NP, even though the former hardly displays any variation in its samples, as seen in \cref{fig:synthetic}. The figure also shows how NP and ANP struggle to fit high frequency signals, while \textsc{ConvCNP} and \textsc{GP-ConvCNP} are able to. The sample diversity in \textsc{GP-ConvCNP} is larger than in ANP, but samples are only significantly different from the mean prediction when further away from the context points in areas of high predictive uncertainty (shaded areas correspond to $1\sigma$). In contrast, samples from the NP are more diverse throughout, at the expense of accurately matching the context points.

\begin{table}[h]
    \caption{Results for the image experiments, in terms of predictive log-likelihood (i.e. higher is better) on the respective test sets. Errors represent 1 standard deviation over 10 runs with different seeds. Bold indicates a significant difference, i.e. when the difference is larger than the root sum of squares of the standard deviations. \textsc{GP-ConvCNP} outperforms \textsc{ConvCNP} overall, with a slight (non-significant) advantage for \textsc{ConvCNP} on MNIST. Visual examples and more details on the image experiments are given in \cref{sec:appendix:imageexperiments}.}
    \label{tab:image}
    \centering
    \sisetup{detect-weight=true,detect-inline-weight=math}
    \begin{tabular}{lS[table-format=2.3(3)]S[table-format=1.3(3)]}
    \toprule
    & \multicolumn{1}{c}{\textsc{ConvCNP}} & \multicolumn{1}{c}{\textsc{GP-ConvCNP}} \\
    \midrule
    MNIST & 4.133(57) & 4.077(26) \\
    CIFAR10 & 2.462(6) & \bfseries 2.744(8) \\
    CelebA & 2.212(6) & \bfseries 2.468(8) \\
    \bottomrule
    \end{tabular}
\end{table}

Examples from the temperature time series dataset can be seen in \cref{fig:temperature}. The key characteristic of the signal is the temperature change between day and night, making it a high frequency signal not unlike the weakly periodic GP samples in the synthetic dataset. NP and ANP were not able to fit these signals, as can be seen in \cref{fig:appendix:temperature_np}. The top row of \cref{fig:temperature} shows an example of the regular interpolation task, the bottom row an example of the extrapolation task, which we deem an important aspect of generalization. \textsc{ConvCNP} and \textsc{GP-ConvCNP} are both able to interpolate as well as extrapolate the correct temperature pattern, but occasionally \textsc{ConvCNP} underestimates the amplitude when extrapolating. We also show an example of a periodic GP using an Exponential Sine-Squared kernel, which is a common choice for periodic signals. It fails to capture finer variations in the signal and often struggles to infer the right frequency, which results in its poor extrapolation performance in \cref{tab:real}. We find that while \textsc{ConvCNP} and \textsc{GP-ConvCNP} perform on par for the interpolation task, \textsc{GP-ConvCNP} performs significantly better than the other methods on the extrapolation task.

To measure how robust the different members of the Neural Process family are to a distribution shift at test time, we train models on population dynamics simulated as Lotka-Volterra processes, and evaluate performance both on simulated (\emph{in-distribution}) and real world (\emph{out-of-distribution}) data. The real world dataset, along with a simulated example, can be seen in \cref{fig:population}. While both \textsc{ConvCNP} and \textsc{GP-ConvCNP} fit the simulated data well, they struggle with the test interval on the real data. This is reflected in \cref{tab:real} as well, where we find that \textsc{ConvCNP} performs better than \textsc{GP-ConvCNP} (even significantly so, albeit not with a huge difference) on the simulated data. Applied to the real world dataset, all methods experience a large drop in performance, indicating that this is indeed a significant distribution shift. \textsc{GP-ConvCNP} is by far the best performing method here, which is likely because of a better estimate of the preditive uncertainty. Note how the uncertainty predicted by \textsc{ConvCNP} is smaller than that of \textsc{GP-ConvCNP} in \cref{fig:population} (the figure shows $1\sigma$). The predictions we show here are from the best performing seed in each case, other \textsc{ConvCNP} models predicted an even narrower distribution. We selected this particular interval for testing because it's the same interval \cite{gordon_convolutional_2019} show in the \textsc{ConvCNP} paper. We also evaluated with context points drawn randomly from the entire interval (i.e. the same way we evaluate on the simulated data), and \textsc{GP-ConvCNP} still performs significantly better than the competing approaches (see \cref{tab:appendix:population}).

\textsc{ConvCNP} also showed performance improvements compared to NP and ANP when applied to image data. While the focus of our work is on time series, we were also interested to see if our model yields any benefits in this domain. It does indeed, as seen in \cref{tab:image}, where \textsc{GP-ConvCNP} outperforms \textsc{ConvCNP} on both CIFAR10 and CelebA (\textsc{ConvCNP} has a non-significant advantage on MNIST). Examples are given in \cref{sec:appendix:imageexperiments}, where we don't see any meaningful difference in visual quality. The latter only \enquote{measures} the quality of the mean prediction, so we suspect that the performance improvement is due to a more accurate predictive uncertainty.

\section{Related Work}
\label{sec:relatedwork}

Neural Processes have inspired a number of works outside of the ones we discuss. \cite{louizos_functional_2019} propose to not merge observations into a global latent space, but instead learn conditional relationships between them. This is especially suitable for semantically meaningful clustering and classification. \cite{singh_sequential_2019} and \cite{willi_recurrent_2019} address the problem of overlapping and changing dynamics in the generating process of the data, a special case we do not include here. With a simple Gaussian kernel, we wouldn't expect our model to perform well in that scenario, but one could of course introduce inductive bias in the form of e.g. non-stationary kernels, when translation equivariance is no longer desired. NPs have also been scaled to extremely complex output spaces like in \emph{Generative Query Networks} \citep{eslami_neural_2018,rosenbaum_learning_2018}, where a single observation is a full image. GQN directly relates to the problem of (3D) scene understanding \citep{sitzmann_scene_2019,engelcke_genesis:_2019}. 

\cite{gordon_convolutional_2019} build their work (\textsc{ConvCNP}) upon recent contributions in the area of \emph{learning on sets}, i.e. neural networks with set-valued inputs \citep{zaheer_deep_2017,wagstaff_limitations_2019}, which has mostly been explored in the context of point clouds \citep{qi_pointnet_2017,charles_pointnet_2017,wu_pointconv_2019}. Especially the work of \cite{wu_pointconv_2019} is closely related to \cite{gordon_convolutional_2019}, also employing a CNN on a kernel density estimate, but their application is not concerned with time series. \emph{Bayesian Neural Networks} \citep{neal_bayesian_1996,graves_practical_2011,hernandez-lobato_probabilistic_2015} also address the problem of learning distributions over functions, but often implicitly, in the sense that the distributions over the weights are used to estimate uncertainty \citep{blundell_weight_2015,gal_dropout_2016}. We are interested in this too, but in our scenario we want to be able to condition on observations at test time.

The main limitation of \emph{Gaussian Processes} is their computational complexity and many works are dedicated to improving this aspect, often via approximations based on inducing points \citep{snelson_sparse_2006,titsias_variational_2009,gardner_product_2018,wilson_kernel_2015} but also other approaches \citep{deisenroth_distributed_2015,rahimi_random_2008,le_fastfood_2013,cheng_variational_2017,hensman_gaussian_2013,hensman_scalable_2015,salimbeni_orthogonally_2018}, even for exact GPs \citep{wang_exact_2019}. Rather than competing with these approaches, our model will be able to leverage developments in this area. Some of the above try to find more efficient kernel representations and are thus closely related to the idea of \emph{kernel learning}, i.e. the idea to combine the expressiveness of (deep) learning approaches with the flexibility of kernel methods, for example \cite{yang_carte_2015,wilson_deep_2015,wilson_stochastic_2016,tossou_adaptive_2019,calandra_manifold_2016}. The key difference to our work is that these approaches attempt to learn kernels as an input to a kernel method, while we learn to make the output of a kernel method more expressive.

\section{Discussion}
\label{sec:discussion}

We have presented a new model in the Neural Process family that extends \textsc{ConvCNP} by incorporating a Gaussian Process into it. We show on both synthetic and real time series that this improves performance overall, but most markedly when generalization is required: our model, \textsc{GP-ConvCNP}, can better extrapolate to regions far from the provided context points and is more robust when moving to real world data after training on simulated data. We further retain translation equivariance, a key feature of \textsc{ConvCNP}, as long as a stationary kernel is used for the GP. The introduction of the latter also allows us to draw multiple samples from the model, where the distribution of samples from our model better matches the samples from an oracle than those from a regular Neural Process or an Attentive Neural Process do. Our model uses the prediction from a GP with an EQ-kernel as an initial estimate. Interestingly, this estimate needn't be very good---our model can learn periodicity even with a non-periodic input kernel---but when it is, our model can fully leverage it and even match the performance of an oracle, as seen in \cref{tab:synthetic}. An advantage all Neural Process flavors enjoy compared to many conventional time series prediction methods such as ARIMA models (see e.g. \cite{hyndman_forecasting_2018}) is that they naturally work on non-uniform time series, with observations acquired at arbitrary times.

Of course, with the benefits of GPs we also inherit their limitations. GPs are typically slow, naively requiring $O(N^3)$ operations in the number of context observations, and our model inherits this complexity. While this was a non-issue on the time series data used in our work, \textsc{GP-ConCNP} was noticably slower than \textsc{ConvCNP} (roughly 1.5x) in the image experiments, which we included for a more complete comparison with \textsc{ConvCNP}. Our model still outperformed \textsc{ConvCNP}, but for larger images the improved performance will likely not be worth the additional cost. Making GPs faster is a very active research area, as outlined above. For our model specifically it seems reasonable to leverage work on deep kernels \citep{wilson_deep_2015} or to learn mappings before the GP prediction like in \cite{calandra_manifold_2016} in order to learn more meaningful GP posteriors that capture information about the training distribution. We do expect that our model is well suited to also work with these approximate methods, as we modify the prediction from the GP with a powerful neural network that should be able to correct minor approximation errors. For example, KISS-GP \cite{wilson_kernel_2015} only has linear complexity, so incorporating it or one of the many other efficient approximate methods into our model should allow it to scale to much larger datasets. We leave a verification of this for future work.



\bibliography{petersen_355}

\begin{thebibliography}{49}
\providecommand{\natexlab}[1]{#1}
\providecommand{\url}[1]{\texttt{#1}}
\expandafter\ifx\csname urlstyle\endcsname\relax
  \providecommand{\doi}[1]{doi: #1}\else
  \providecommand{\doi}{doi: \begingroup \urlstyle{rm}\Url}\fi

\bibitem[Bishop(1995)]{bishop_1995}
Christopher~M. Bishop.
\newblock \emph{Neural Networks for Pattern Recognition}.
\newblock Oxford University Press, Inc., 1995.

\bibitem[Blundell et~al.(2015)Blundell, Cornebise, Kavukcuoglu, and
  Wierstra]{blundell_weight_2015}
Charles Blundell, Julien Cornebise, Koray Kavukcuoglu, and Daan Wierstra.
\newblock Weight uncertainty in neural networks.
\newblock In \emph{International Conference on Machine Learning}, pages
  1613--1622, 2015.

\bibitem[Calandra et~al.(2016)Calandra, Peters, Rasmussen, and
  Deisenroth]{calandra_manifold_2016}
Roberto Calandra, Jan Peters, Carl~Edward Rasmussen, and Marc~Peter Deisenroth.
\newblock Manifold gaussian processes for regression.
\newblock \emph{{arXiv}:1402.5876 [cs, stat]}, 2016.

\bibitem[Cheng and Boots(2017)]{cheng_variational_2017}
Ching-An Cheng and Byron Boots.
\newblock Variational inference for gaussian process models with linear
  complexity.
\newblock In \emph{Advances in Neural Information Processing Systems 30}, pages
  5184--5194. 2017.

\bibitem[Deisenroth and Ng(2015)]{deisenroth_distributed_2015}
Marc Deisenroth and Jun~Wei Ng.
\newblock Distributed gaussian processes.
\newblock In \emph{International Conference on Machine Learning}, pages
  1481--1490, 2015.

\bibitem[Engelcke et~al.(2020)Engelcke, Kosiorek, Jones, and
  Posner]{engelcke_genesis:_2019}
Martin Engelcke, Adam~R. Kosiorek, Oiwi~Parker Jones, and Ingmar Posner.
\newblock {GENESIS}: Generative scene inference and sampling with
  object-centric latent representations.
\newblock In \emph{International Conference on Learning Representations}, 2020.

\bibitem[Eslami et~al.(2018)Eslami, Rezende, Besse, Viola, Morcos, Garnelo,
  Ruderman, Rusu, Danihelka, Gregor, Reichert, Buesing, Weber, Vinyals,
  Rosenbaum, Rabinowitz, King, Hillier, Botvinick, Wierstra, Kavukcuoglu, and
  Hassabis]{eslami_neural_2018}
S.~M.~Ali Eslami, Danilo~Jimenez Rezende, Frederic Besse, Fabio Viola, Ari~S.
  Morcos, Marta Garnelo, Avraham Ruderman, Andrei~A. Rusu, Ivo Danihelka, Karol
  Gregor, David~P. Reichert, Lars Buesing, Theophane Weber, Oriol Vinyals, Dan
  Rosenbaum, Neil Rabinowitz, Helen King, Chloe Hillier, Matt Botvinick, Daan
  Wierstra, Koray Kavukcuoglu, and Demis Hassabis.
\newblock Neural scene representation and rendering.
\newblock \emph{Science}, 360\penalty0 (6394):\penalty0 1204--1210, 2018.

\bibitem[Gal and Ghahramani(2016)]{gal_dropout_2016}
Yarin Gal and Zoubin Ghahramani.
\newblock Dropout as a bayesian approximation: Representing model uncertainty
  in deep learning.
\newblock In \emph{International Conference on Machine Learning}, pages
  1050--1059, 2016.

\bibitem[Gardner et~al.(2018)Gardner, Pleiss, Wu, Weinberger, and
  Wilson]{gardner_product_2018}
Jacob Gardner, Geoff Pleiss, Ruihan Wu, Kilian Weinberger, and Andrew Wilson.
\newblock Product kernel interpolation for scalable gaussian processes.
\newblock In \emph{International Conference on Artificial Intelligence and
  Statistics}, pages 1407--1416, 2018.

\bibitem[Garnelo et~al.(2018{\natexlab{a}})Garnelo, Rosenbaum, Maddison,
  Ramalho, Saxton, Shanahan, Teh, Rezende, and
  Eslami]{garnelo_conditional_2018}
Marta Garnelo, Dan Rosenbaum, Christopher Maddison, Tiago Ramalho, David
  Saxton, Murray Shanahan, Yee~Whye Teh, Danilo Rezende, and S.~M.~Ali Eslami.
\newblock Conditional neural processes.
\newblock In \emph{International Conference on Machine Learning}, pages
  1704--1713, 2018{\natexlab{a}}.

\bibitem[Garnelo et~al.(2018{\natexlab{b}})Garnelo, Schwarz, Rosenbaum, Viola,
  Rezende, Eslami, and Teh]{garnelo_neural_2018}
Marta Garnelo, Jonathan Schwarz, Dan Rosenbaum, Fabio Viola, Danilo~J. Rezende,
  S.~M.~Ali Eslami, and Yee~Whye Teh.
\newblock Neural processes.
\newblock In \emph{ICML Workshop on Theoretical Foundations and Applications of
  Deep Generative Models}, 2018{\natexlab{b}}.

\bibitem[Gordon et~al.(2020)Gordon, Bruinsma, Foong, Requeima, Dubois, and
  Turner]{gordon_convolutional_2019}
Jonathan Gordon, Wessel~P. Bruinsma, Andrew Y.~K. Foong, James Requeima, Yann
  Dubois, and Richard~E. Turner.
\newblock Convolutional conditional neural processes.
\newblock In \emph{International Conference on Learning Representations}, 2020.

\bibitem[Graves(2011)]{graves_practical_2011}
Alex Graves.
\newblock Practical variational inference for neural networks.
\newblock In \emph{Advances in Neural Information Processing Systems 24}, pages
  2348--2356. 2011.

\bibitem[Hensman et~al.(2013)Hensman, Fusi, and
  Lawrence]{hensman_gaussian_2013}
James Hensman, Nicolò Fusi, and Neil~D. Lawrence.
\newblock Gaussian processes for big data.
\newblock In \emph{Proceedings of the Twenty-Ninth Conference on Uncertainty in
  Artificial Intelligence}, pages 282--290, 2013.

\bibitem[Hensman et~al.(2015)Hensman, Matthews, and
  Ghahramani]{hensman_scalable_2015}
James Hensman, Alexander Matthews, and Zoubin Ghahramani.
\newblock Scalable variational gaussian process classification.
\newblock In \emph{Internation Conference on Artificial Intelligence and
  Statistics}, pages 351--360, 2015.

\bibitem[Hernández-Lobato and
  Adams(2015)]{hernandez-lobato_probabilistic_2015}
José~Miguel Hernández-Lobato and Ryan~P. Adams.
\newblock Probabilistic backpropagation for scalable learning of bayesian
  neural networks.
\newblock In \emph{International Conference on Machine Learning}, volume~37,
  pages 1861--1869, 2015.

\bibitem[Hyndman and Athanasopoulos(2018)]{hyndman_forecasting_2018}
Rob~J. Hyndman and George Athanasopoulos.
\newblock \emph{Forecasting: principles and practice}.
\newblock {OTexts}, 2018.

\bibitem[Kim et~al.(2019)Kim, Mnih, Schwarz, Garnelo, Eslami, Rosenbaum,
  Vinyals, and Teh]{kim_attentive_2019}
Hyunjik Kim, Andriy Mnih, Jonathan Schwarz, Marta Garnelo, Ali Eslami, Dan
  Rosenbaum, Oriol Vinyals, and Yee~Whye Teh.
\newblock Attentive neural processes.
\newblock In \emph{International Conference on Learning Representations}, 2019.

\bibitem[Krizhevsky(2009)]{krizhevsky_learning-multiple_2009}
Alex Krizhevsky.
\newblock Learning multiple layers of features from tiny images.
\newblock Technical report, 2009.

\bibitem[Le et~al.(2013)Le, Sarlos, and Smola]{le_fastfood_2013}
Quoc Le, Tamas Sarlos, and Alexander Smola.
\newblock Fastfood - computing hilbert space expansions in loglinear time.
\newblock In \emph{International Conference on Machine Learning}, pages
  244--252, 2013.

\bibitem[Le et~al.(2018)Le, Kim, Garnelo, Rosenbaum, Schwarz, and
  Teh]{le_empirical_2018}
Tuan~Anh Le, Hyunjik Kim, Marta Garnelo, Dan Rosenbaum, Jonathan Schwarz, and
  Yee~Whye Teh.
\newblock Empirical evaluation of neural process objectives.
\newblock In \emph{{NeurIPS} Bayesian Deep Learning Workshop}, 2018.

\bibitem[Lecun et~al.()Lecun, Bottou, Bengio, and
  Haffner]{lecun_gradient-based_1998}
Y.~Lecun, L.~Bottou, Y.~Bengio, and P.~Haffner.
\newblock Gradient-based learning applied to document recognition.
\newblock 86\penalty0 (11):\penalty0 2278--2324.

\bibitem[Leigh(1968)]{leigh_1968}
Egbert~G Leigh.
\newblock \emph{Ecological role of Volterra's equations}.
\newblock Lectures on mathematics in the life sciences. Princeton University,
  1968.

\bibitem[Liu et~al.(2015)Liu, Luo, Wang, and Tang]{liufaceattributes_2015}
Ziwei Liu, Ping Luo, Xiaogang Wang, and Xiaoou Tang.
\newblock Deep learning face attributes in the wild.
\newblock In \emph{Proceedings of International Conference on Computer Vision
  (ICCV)}, December 2015.

\bibitem[Louizos et~al.(2019)Louizos, Shi, Schutte, and
  Welling]{louizos_functional_2019}
Christos Louizos, Xiahan Shi, Klamer Schutte, and Max Welling.
\newblock The functional neural process.
\newblock In \emph{Advances in Neural Information Processing Systems 32}, pages
  8743--8754. 2019.

\bibitem[Neal(1996)]{neal_bayesian_1996}
Radford~M. Neal.
\newblock \emph{Bayesian Learning for Neural Networks}.
\newblock Lecture Notes in Statistics. Springer, 1996.

\bibitem[Qi et~al.(2017{\natexlab{a}})Qi, Su, Kaichun, and
  Guibas]{charles_pointnet_2017}
Charles~R. Qi, Hao Su, Mo~Kaichun, and Leonidas~J. Guibas.
\newblock {PointNet}: Deep learning on point sets for 3d classification and
  segmentation.
\newblock In \emph{{IEEE} Conference on Computer Vision and Pattern
  Recognition}, pages 77--85, 2017{\natexlab{a}}.

\bibitem[Qi et~al.(2017{\natexlab{b}})Qi, Yi, Su, and Guibas]{qi_pointnet_2017}
Charles~Ruizhongtai Qi, Li~Yi, Hao Su, and Leonidas~J Guibas.
\newblock {PointNet}++: Deep hierarchical feature learning on point sets in a
  metric space.
\newblock In \emph{Advances in Neural Information Processing Systems 30}, pages
  5099--5108. 2017{\natexlab{b}}.

\bibitem[Rahimi and Recht(2007)]{rahimi_random_2008}
Ali Rahimi and Benjamin Recht.
\newblock Random features for large-scale kernel machines.
\newblock In \emph{Advances in Neural Information Processing Systems 20}, pages
  1177--1184. 2007.

\bibitem[Rasmussen and Williams(2006)]{rasmussen_gaussian_2006}
Carl~Edward Rasmussen and C.~K.~I. Williams.
\newblock \emph{Gaussian Processes for Machine Learning}.
\newblock {MIT} Press, 2006.

\bibitem[Roberts et~al.(2013)Roberts, Osborne, Ebden, Reece, Gibson, and
  Aigrain]{roberts_gaussian_2013}
S.~Roberts, M.~Osborne, M.~Ebden, S.~Reece, N.~Gibson, and S.~Aigrain.
\newblock Gaussian processes for time-series modelling.
\newblock 371\penalty0 (1984):\penalty0 20110550, 2013.

\bibitem[Rosenbaum et~al.(2018)Rosenbaum, Besse, Viola, Rezende, and
  Eslami]{rosenbaum_learning_2018}
Dan Rosenbaum, Frederic Besse, Fabio Viola, Danilo~J. Rezende, and S.~M.~Ali
  Eslami.
\newblock Learning models for visual 3d localization with implicit mapping.
\newblock In \emph{{NeurIPS} Bayesian Deep Learning Workshop}, 2018.

\bibitem[Salimbeni et~al.(2018)Salimbeni, Cheng, Boots, and
  Deisenroth]{salimbeni_orthogonally_2018}
Hugh Salimbeni, Ching-An Cheng, Byron Boots, and Marc Deisenroth.
\newblock Orthogonally decoupled variational gaussian processes.
\newblock In \emph{Advances in Neural Information Processing Systems 31}, pages
  8711--8720. 2018.

\bibitem[Singh et~al.(2019)Singh, Yoon, Son, and Ahn]{singh_sequential_2019}
Gautam Singh, Jaesik Yoon, Youngsung Son, and Sungjin Ahn.
\newblock Sequential neural processes.
\newblock In \emph{Advances in Neural Information Processing Systems 32}, pages
  10254--10264. 2019.

\bibitem[Sitzmann et~al.(2019)Sitzmann, Zollhoefer, and
  Wetzstein]{sitzmann_scene_2019}
Vincent Sitzmann, Michael Zollhoefer, and Gordon Wetzstein.
\newblock Scene representation networks: Continuous 3d-structure-aware neural
  scene representations.
\newblock In \emph{Advances in Neural Information Processing Systems 32}, pages
  1119--1130. 2019.

\bibitem[Snelson and Ghahramani(2006)]{snelson_sparse_2006}
Edward Snelson and Zoubin Ghahramani.
\newblock Sparse gaussian processes using pseudo-inputs.
\newblock In \emph{Advances in Neural Information Processing Systems 18}, pages
  1257--1264. 2006.

\bibitem[Sriperumbudur et~al.(2009)Sriperumbudur, Fukumizu, Gretton,
  Schölkopf, and Lanckriet]{sriperumbudur_integral_2009}
Bharath~K. Sriperumbudur, Kenji Fukumizu, Arthur Gretton, Bernhard Schölkopf,
  and Gert R.~G. Lanckriet.
\newblock On integral probability metrics, phi-divergences and binary
  classification.
\newblock \emph{{arXiv}:0901.2698 [cs, math]}, 2009.

\bibitem[Titsias(2009)]{titsias_variational_2009}
Michalis Titsias.
\newblock Variational learning of inducing variables in sparse gaussian
  processes.
\newblock In \emph{International Conference on Artificial Intelligence and
  Statistics}, pages 567--574, 2009.

\bibitem[Tossou et~al.(2019)Tossou, Dura, Laviolette, Marchand, and
  Lacoste]{tossou_adaptive_2019}
Prudencio Tossou, Basile Dura, Francois Laviolette, Mario Marchand, and
  Alexandre Lacoste.
\newblock Adaptive deep kernel learning.
\newblock \emph{{arXiv}:1905.12131 [cs, stat]}, 2019.

\bibitem[Volpi et~al.(2018)Volpi, Namkoong, Sener, Duchi, Murino, and
  Savarese]{volpi_2018}
Riccardo Volpi, Hongseok Namkoong, Ozan Sener, John Duchi, Vittorio Murino, and
  Silvio Savarese.
\newblock Generalizing to unseen domains via adversarial data augmentation.
\newblock In \emph{Advances in Neural Information Processing Systems 31}, page
  5339–5349, 2018.

\bibitem[Wagstaff et~al.(2019)Wagstaff, Fuchs, Engelcke, Posner, and
  Osborne]{wagstaff_limitations_2019}
Edward Wagstaff, Fabian~B. Fuchs, Martin Engelcke, Ingmar Posner, and Michael
  Osborne.
\newblock On the limitations of representing functions on sets.
\newblock In \emph{International Conference on Machine Learning}, 2019.

\bibitem[Wang et~al.(2019)Wang, Pleiss, Gardner, Tyree, Weinberger, and
  Wilson]{wang_exact_2019}
Ke~Wang, Geoff Pleiss, Jacob Gardner, Stephen Tyree, Kilian~Q Weinberger, and
  Andrew~Gordon Wilson.
\newblock Exact gaussian processes on a million data points.
\newblock In \emph{Advances in Neural Information Processing Systems 32}, pages
  14622--14632. 2019.

\bibitem[Willi et~al.(2019)Willi, Masci, Schmidhuber, and
  Osendorfer]{willi_recurrent_2019}
Timon Willi, Jonathan Masci, Jürgen Schmidhuber, and Christian Osendorfer.
\newblock Recurrent neural processes.
\newblock \emph{{arXiv}:1906.05915 [cs, stat]}, 2019.

\bibitem[Wilson et~al.(2016{\natexlab{a}})Wilson, Hu, Salakhutdinov, and
  Xing]{wilson_stochastic_2016}
Andrew~G Wilson, Zhiting Hu, Russ~R Salakhutdinov, and Eric~P Xing.
\newblock Stochastic variational deep kernel learning.
\newblock In \emph{Advances in Neural Information Processing Systems 29}, pages
  2586--2594. 2016{\natexlab{a}}.

\bibitem[Wilson and Nickisch(2015)]{wilson_kernel_2015}
Andrew~Gordon Wilson and Hannes Nickisch.
\newblock Kernel interpolation for scalable structured gaussian processes
  ({KISS}-{GP}).
\newblock In \emph{International Conference on Machine Learning}, 2015.

\bibitem[Wilson et~al.(2016{\natexlab{b}})Wilson, Hu, Salakhutdinov, and
  Xing]{wilson_deep_2015}
Andrew~Gordon Wilson, Zhiting Hu, Ruslan Salakhutdinov, and Eric~P. Xing.
\newblock Deep kernel learning.
\newblock In \emph{International Conference on Artificial Intelligence and
  Statistics}, 2016{\natexlab{b}}.

\bibitem[Wu et~al.(2019)Wu, Qi, and Fuxin]{wu_pointconv_2019}
Wenxuan Wu, Zhongang Qi, and Li~Fuxin.
\newblock {PointConv}: Deep convolutional networks on 3d point clouds.
\newblock In \emph{{IEEE} Conference on Computer Vision and Pattern
  Recognition}, pages 9621--9630, 2019.

\bibitem[Yang et~al.(2015)Yang, Wilson, Smola, and Song]{yang_carte_2015}
Zichao Yang, Andrew Wilson, Alex Smola, and Le~Song.
\newblock A la carte – learning fast kernels.
\newblock In \emph{International Conference on Artificial Intelligence and
  Statistics}, pages 1098--1106, 2015.

\bibitem[Zaheer et~al.(2017)Zaheer, Kottur, Ravanbakhsh, Poczos, Salakhutdinov,
  and Smola]{zaheer_deep_2017}
Manzil Zaheer, Satwik Kottur, Siamak Ravanbakhsh, Barnabas Poczos, Russ~R
  Salakhutdinov, and Alexander~J Smola.
\newblock Deep sets.
\newblock In \emph{Advances in Neural Information Processing Systems 30}, pages
  3391--3401. 2017.

\end{thebibliography}

\clearpage
\appendix

\renewcommand\thefigure{A.\arabic{figure}}    
\setcounter{figure}{0}
\renewcommand\thetable{A.\arabic{table}}    
\setcounter{table}{0} 

\begin{figure*}[t]
    \centering
    \includegraphics[width=\textwidth]{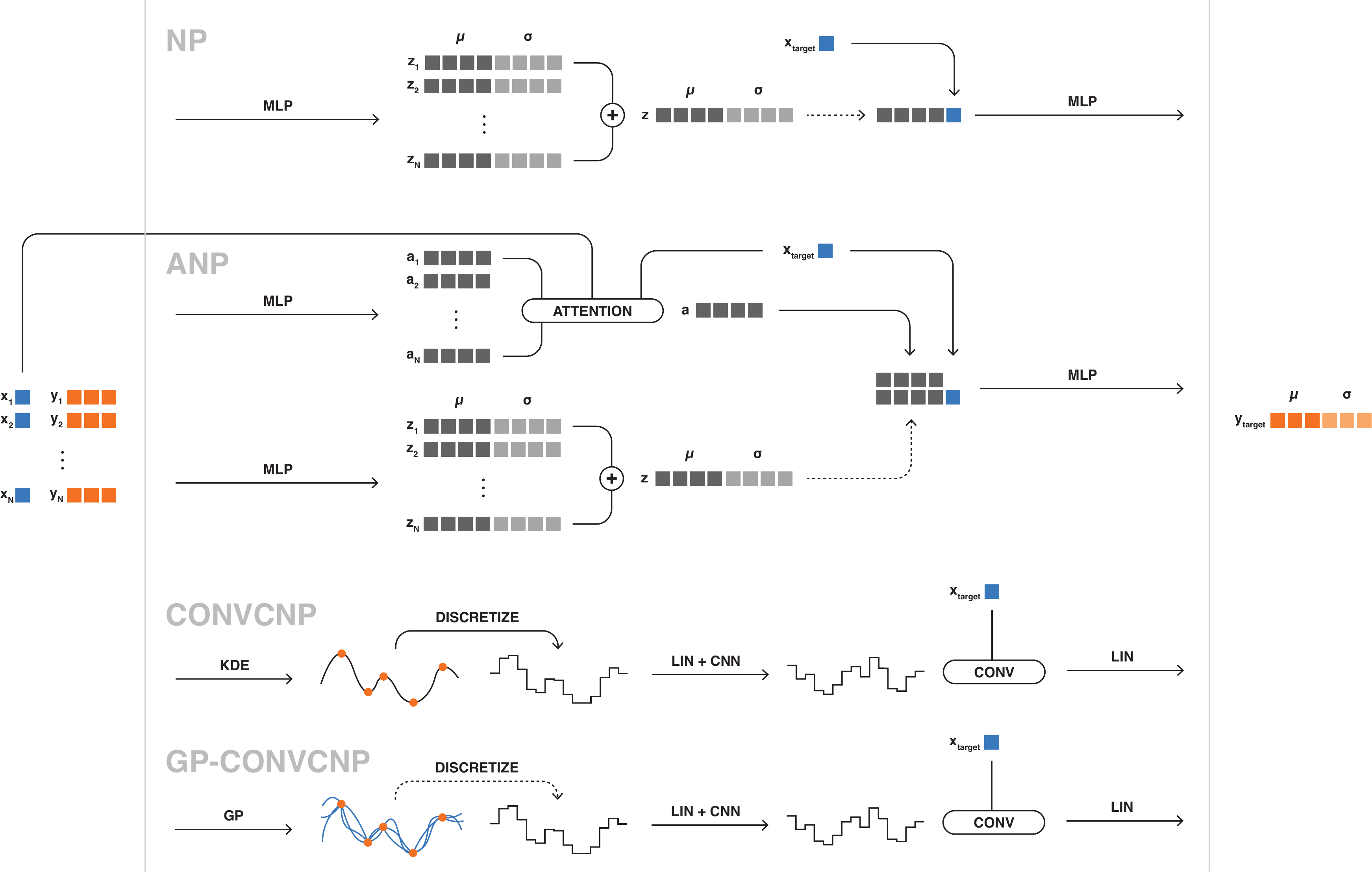}
    \caption{Schematic overview of the different methods used in this work. Dotted lines indicate sampling and we use the following acronyms: multilayer perceptron (MLP), kernel density estimate (KDE), Gaussian Process (GP), linear layer (LIN), convolutional neural network (CNN). (\emph{First row}) Neural Processes (NP) encode each context point $(x_c,y_c)$ into a representation $z_c$. These are then averaged to form a global representation $z$. A sample from the global representation is concatenated with the target input $x_t$ to predict the target output $y_t$. (\emph{Second row}) Attentive Neural Processes (ANP) contain a NP, but have a second deterministic path. In this path, the context pairs are also encoded separately into representations $\mathbf{a_c}$. These are then combined via an atttention mechanism that uses $x_t$ as the query, $\mathbf{x_c}$ as the keys and $\mathbf{a_c}$ as the values. The resulting representation $a$ is concatenated with the representation from the NP path and the target input to predict the target output. (\emph{Third row}) \textsc{ConvCNP} performs a kernel density estimate on the context observations $(\mathbf{x_c},\mathbf{y_c})$, thus mapping to a continuous representation. This representation is evaluated on a grid, i.e. discretized, and a projection and CNN operate on the discretized representation. The result is evaluated at a target input $x_t$ by performing a convolution with the discretized representation and finally projected to predict the target output. (\emph{Fourth row}) \textsc{GP-ConvCNP} works similar to \textsc{ConvCNP}, but instead of a deterministic kernel density estimate a Gaussian Process is applied to the context. We sample from the GP posterior and discretize the result, continuing with the same operations as in \textsc{ConvCNP}. Note that for visual purposes, the KDE and GP outputs are one-dimensional, but in reality the output space can have any number of dimensions.}
    \label{fig:appendix:methods}
\end{figure*}

\clearpage








\section{Method Descriptions}
\label{sec:appendix:methods}

\cref{fig:appendix:methods} shows schematic representations of the different methods used in this work, and a description is given in the figure caption. The MLPs in both NP and ANP have 6 hidden layers with 128 channels each, and the input and output sizes are adjusted to match the dimensions of data and latent representations. The latent representation in both models has 128 dimensions, so that the encoders for the NP and the NP path in ANP have 256 output channels to represent both the mean and the standard deviation of a Gaussian distribution (in practice, we predict the log-variance, not the standard deviation). The attention mechanism in ANP also uses 128 as the embedding dimension. These configurations follow \cite{le_empirical_2018}, who evaluated several different configurations for NP and ANP.

\textsc{ConvCNP} and \textsc{GP-ConvCNP} both use a Gaussian kernel with a learnable length scale $l$ to map the input to a continuous representation, given by

\begin{equation}
    k(x,x')=\exp\left(-\frac{|x-x'|^2}{2l^2}\right)
\end{equation}

The result is discretized onto a grid, which we obtain by taking the minimum and maximum of the target inputs as the value range, padded by 0.1 units. The grid is constructed over this range with a resolution of 20 points per unit. The discretized representations are projected to 8 channels before a CNN is applied. The CNN is a 12-layer residual network with ReLU activations. The number of channels in the convolutional layers doubles every second layer for the first 6 layers and is then decreased symmetrically, leading to 8 output channels. Residual connections are implemented via concatenation. Predictions are obtained by convolving the CNN output with a target input, followed by a final projection.

\section{Optimization}
\label{sec:appendix:optimization}

Recall that our optimization objective is

\begin{equation}
    \max_\theta\sum_{f\in\mathcal{F}}\log p_\theta(\mathbf{y_t}|\mathbf{x_t},\mathbf{x_c},\mathbf{y_c})
\label{eq:appendix:objective}
\end{equation}

which we can rewrite as

\begin{equation}
    \max_\theta\sum_{f\in\mathcal{F}}\log p_\theta(\mathbf{y_t}|\mathbf{x_t},Z)
\end{equation}

where $Z$ is given by the different $E$ that \emph{encode} the context defined in \cref{sec:methods}. For \textsc{ConvCNP} this is deterministic, so we can maximize \cref{eq:appendix:objective} directly. For the other methods we can again rewrite the summands as

\begin{equation}
    \log p(\mathbf{y_t}|\mathbf{x_t},\mathbf{x_c},\mathbf{y_c})=\log\mathop{\mathbb{E}}_{z\sim p(z|\mathbf{x_c},\mathbf{y_c})} p(\mathbf{y_t}|\mathbf{x_t},z)
\end{equation}

where we now distinguish $z$ as an expression of $Z$. In \textsc{GP-ConvCNP}, $p(z|\mathbf{x_c},\mathbf{y_c})$ is given by the GP posterior, so for training we would need to integrate over the posterior. In practice, we just draw a single sample, which is common practice in stochastic mini-batch training. Approximating the expectation with this sample, we can also directly maximize the log-likelihood.

In contrast to the above, $p(z|\mathbf{x_c},\mathbf{y_c})$ is an unknown or intractable mapping in NP and ANP, so we employ variational inference, i.e. we approximate $p(z|\mathbf{x_c},\mathbf{y_c})$ with a member of some family $Q$ that we can find by optimization. The log-likelihood then becomes

\begin{align}
    \log p(\mathbf{y_t}|\mathbf{x_t},\mathbf{x_c},\mathbf{y_c})\geq&\mathop{\mathbb{E}}_{z\sim q(z|\mathbf{x_t},\mathbf{y_t})}\log p(\mathbf{y_t}|\mathbf{x_t},z)\nonumber\\
    &-D_{KL}\left(q(z|\mathbf{x_t},\mathbf{y_t})||p(z|\mathbf{x_c},\mathbf{y_c})\right)\\
    \approx&\mathop{\mathbb{E}}_{z\sim q(z|\mathbf{x_t},\mathbf{y_t})}\log p(\mathbf{y_t}|\mathbf{x_t},z)\nonumber\\
    &-D_{KL}\left(q(z|\mathbf{x_t},\mathbf{y_t})||q(z|\mathbf{x_c},\mathbf{y_c})\right)
    \label{eq:appendix:objectivenpanp}
\end{align}

where the inequality follows from Jensen's inequality. To maximize the LHS it is sufficient to maximize the RHS, and \cref{eq:appendix:objectivenpanp} is what is being optimized in NP and ANP. $q$ corresponds to what we designated as $E$ in \cref{sec:methods}. Like for \textsc{GP-ConvCNP}, we approximate the expectation with a single sample during training.

In our implementation, we use Adam
with an initial learning rate of $0.001$. We train each model for \SI{600000} batches with a batch size of 256. We repeatedly multiply the learning rate by $\gamma=0.995$ after training for 1000 batches.

\section{Data \&\ Evaluation Details}
\label{sec:appendix:dataeval}

\subsection{Synthetic Data}
\label{sub:appendix:synthetic}

For all synthetic time series draws we define the x-axis to cover the interval $[-3,3]$. As outlined in Section 3.2, we draw $N$ context points randomly from this interval, with $N$ a random integer from the range $[3,100)$. We then draw $M$ target points in the same manner, with $M$ a random integer from $[N, 100)$. During training, we add the context points to the target set so that the methods learn to reconstruct the context. These are the different types we evaluate:

\begin{enumerate}
    \item Samples from a Gaussian Process with a Matern-5/2 kernel with lengthscale parameter $l=0.5$. The kernel is given by
     \begin{alignat}{2}
    k(x,x')=&&&\left(1+\frac{\sqrt{5}|x-x'|}{l}+\frac{5|x-x'|^2}{3l^2}\right)\nonumber\\
    &&\cdot\;&\exp\left(-\frac{5|x-x'|}{l}\right)
    \end{alignat}
    \item Samples from a Gaussian Process with a weakly periodic kernel that is given by
    \begin{alignat}{2}
    k(x,x')=&&&\exp\left(-\frac{|x-x'|^2}{8}\right)\nonumber\\
    &&\cdot\;&\exp\Big((\cos(8\pi x)-\cos(8\pi x'))^2\Big)\nonumber\\
    &&\cdot\;&\exp\Big((\sin(8\pi x)-\sin(8\pi x'))^2\Big)
    \end{alignat}
    \item Fourier series that are given by
    \begin{equation}
    f(x)=a_0+\sum_{k=1}^Ka_k\cos(kx-\phi_k)
    \end{equation}
    where $K$ is a random integer from $[10, 20)$ and $a_k$ (including $a_0$) as well as $\phi_k$ are random real numbers drawn from $[-1,1]$.
    \item Step functions, where we draw $S$ stepping points along the x-axis, with $S$ a random integer from $[3,10)$. The interval between two stepping points is assigned a constant value that is drawn from $[-3,3]$. We ensure that each interval is at least $0.1$ units wide and that the step difference is also at least $0.1$ units in magnitude.
\end{enumerate}

\subsection{Temperature Time Series}
\label{sub:appendix:temperature}

The temperature dataset we work with is taken from \url{https://www.kaggle.com/selfishgene/historical-hourly-weather-data}. It consists of hourly temperature measurements in 30 US and Canadian cities as well as 6 Israeli cities, taken continuously over the course of $\sim$5 years. Occasionally there are NaN values reported in the dataset, we either crop those when at the begging/end of a sequence or fill them via linear interpolation. We use the US/Canadian cities as our training and validation set and the Israeli cities as our test set. For both training and testing we draw random sequences of length 720 (i.e. 30 days) from the corresponding set, and then draw $N$ context points and $M$ target points from the sequence, with $N$ from the interval $[20, 100)$ and $M$ from $[N, 100)$. The temperatures for each city are normalized by their respective means and standard deviations, and we define the time range for a given sequence to be $[0, 3]$, so that one time unit is equivalent to 10 days. We evaluate each seed for a model with 100 random samples and report the mean and standard deviation over 5 seeds for each model. For convenience, we include the data with our implementation.

\subsection{Population Dynamics}
\label{sub:appendix:population}

We simulate population dynamics of a predator-prey population with a Lotka-Volterra model. Let $X$ be the number of predators at a given time and $Y$ the number of prey. We draw initial numbers $X$ from $[50, 100)$ and $Y$ from $[100, 150)$. We then draw time increments from an exponential distribution and after each time increment one of the following events occurs:

\begin{enumerate}
    \item A single predator is born with probability proportional to the rate $\theta_0\cdot X\cdot Y$
    \item A single predator dies with probability proportional to the rate $\theta_1\cdot X$
    \item A single prey is born with probability proportional to the rate $\theta_2\cdot Y$
    \item A single prey dies with probability proportional to the rate $\theta_3\cdot X\cdot Y$
\end{enumerate}

The rate of the exponential distribution we draw time increments from is the sum of the above rates. Each population is simulated for 10000 events, and we reject populations that have died out, populations that exceed a total number of 500 individuals at any given point, as well as those where the accumulated time is larger than 100 units. To get value ranges that are better suitable for training, we rescale the time axis by a factor $0.1$ and the population axis by a factor $0.01$. For each population we draw $\theta_0$ from $[0.005, 0.01]$, $\theta_1$ from $[0.5, 0.8]$, $\theta_2$ from $[0.5, 0.8]$ and $\theta_3$ from $[0.005, 0.01]$. These parameters result in roughly 2/3 of the simulated populations matching our criteria. We also tried the parameters reported in \cite{gordon_convolutional_2019}, but found that we had to reject more than $90\%$ of populations, which meant an unreasonably long training time, as the simulation process for the populations is difficult to parallelize and thus rather slow. The $N$ context points and $M$ target points are again drawn randomly from a population, with $N$ from $[20, 100)$ and $M$ from $[\max(70,N),150)$. 

We evaluate models trained on simulated data on real world measurements of a lynx-hare population. The data were recorded at the end of the 19th and the start of the 20th century by the Hudson's Bay Company. To the best of our knowledge, the data represent recorded trades of pelts from the two animals and not direct measurements of the populations. There is no unique source for the data in a tabular format, but we used \url{https://github.com/stan-dev/example-models/blob/master/knitr/lotka-volterra/hudson-bay-lynx-hare} and include the data with our code for convenience. For evaluation, we normalize the data so that the mean population matches the mean of populations in the simulated data and the time interval matches the mean duration of a simulated population.

\subsection{Wasserstein Distance}
\label{sub:appendix:wasserstein}

As outlined in \cref{sec:experiments}, we seek to compare the distribution of samples from a model with a reference distribution, which we have access to for the synthetic examples sampled from a GP in the form of the prediction from the same GP. Comparing distributions is usually done with either some form of \emph{$f$-divergence} 
(e.g. the Kullback-Leibler divergence) or with an \emph{Integral Probability Measure} (IPM)
f-divergences require evaluations of likelihoods in both distributions, while we can only evaluate those under the GP posterior but not in our models. IPM only compare samples from the distributions and are thus suited for our scenario. One of the more well-known measures from this group is the Wasserstein distance
given by:

\begin{equation}
    W_p(P,Q)=\min_\pi\left(\sum_{i=1}^{|P|}||x_i-y_{\pi(i)}||^p\right)^{1/p}
\end{equation}

where $P=\{x_i\}_i$ and $Q=\{y_i\}_i$ are collections of samples from the two distributions. In colloquial terms, the Wasserstein distance is the minimum overall distance between sample pairs, taken over all possible pairings between samples from the two distributions. For this reason the Wasserstein-1 distance is also called the \emph{Earth Mover Distance}. $p$ is the only hyperparameter we need to select, making this measure a very convenient choice. We set $p=2$ so that the underlying distance metric becomes the Euclidean distance.

\section{Additional Results}
\label{sec:appendix:results}

\begin{table}
    \caption{This table corresponds to the rightmost column in \cref{tab:real}, i.e. it shows results on the real world population dynamics data. In \cref{tab:real}, the evaluation was performed as seen in \cref{fig:population}, meaning one contiguous interval on the data was selected as the target region and the rest of the data is provided as context, following \cite{gordon_convolutional_2019}. Here we instead sample the context and target points randomly from the entire interval, like we do in the other experiments as well. For each seed, we average over 100 random draws and report the standard deviation over 5 seeds as errors. While \textsc{ConvCNP} maintains leading performance in terms of reconstruction error, \textsc{GP-ConvCNP} significantly outperforms the other methods in predictive performance, similar to what we found in \cref{tab:real}. All methods perform worse compared to the evaluation method used in \cref{tab:real}.}
    \label{tab:appendix:population}
    \centering
    \sisetup{detect-weight=true,detect-inline-weight=math}
    \begin{tabular}{lS[table-format=2.3(3)]S[table-format=1.3(3)]}
    \toprule
    & \multicolumn{1}{c}{Predictive LL$\uparrow$} & {Recon. Error$\downarrow$} \\
    \midrule
    NP & -36.735(4137) & 0.952(24) \\
    ANP & -38.717(3572) & 0.718(18) \\
    \textsc{ConvCNP} & -28.762(1958) & \bfseries 0.272(8) \\
    \textsc{GP-ConvCNP} & \bfseries -19.252(1846) & 0.343(20) \\
    \bottomrule
    \end{tabular}
\end{table}

\begin{figure*}
    \centering
    \includegraphics[width=\textwidth]{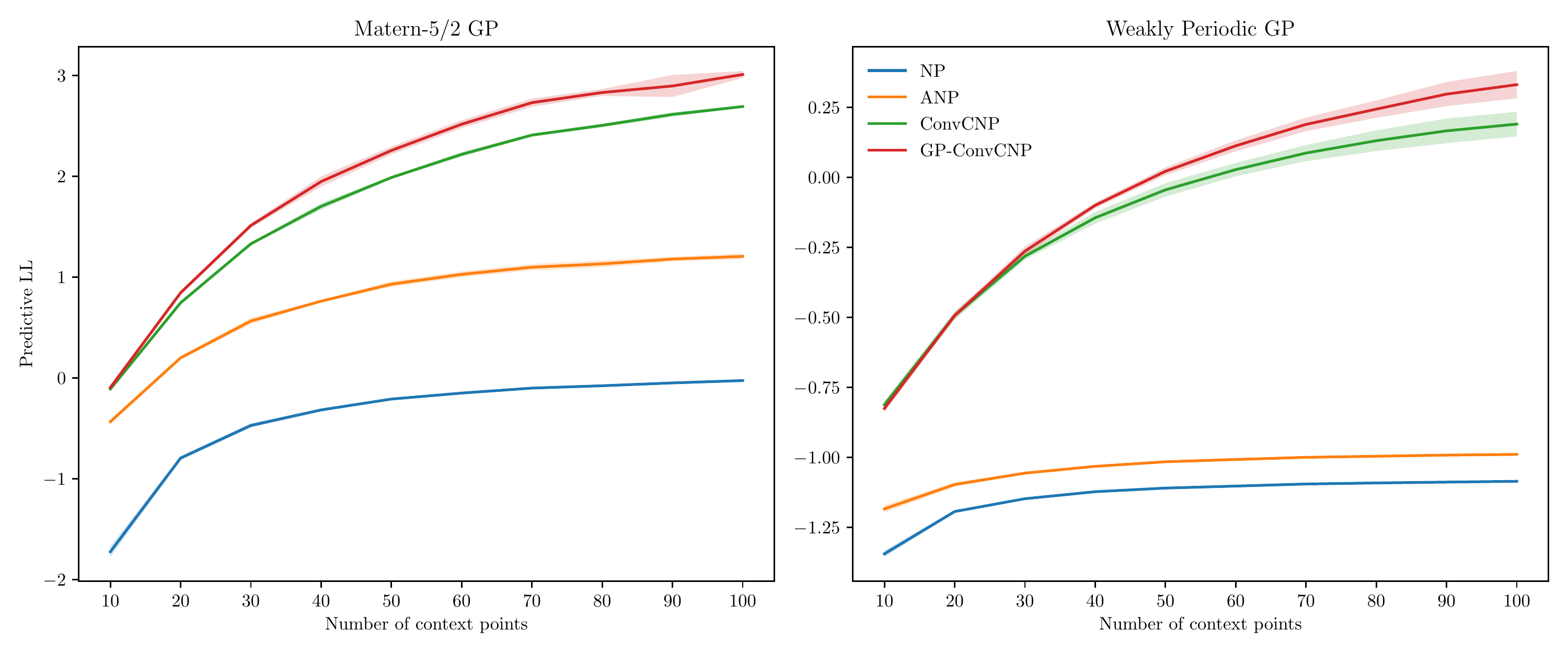}
    \caption{Performance as a function of the number of context points for the two synthetic GP examples. For sparse context points, our model and \textsc{ConvCNP} are on par, while an increasing number of context points leads to an advantage for our model.}
    \label{fig:appendix:synthetic_numcontext}
\end{figure*}

\begin{figure*}
    \centering
    \includegraphics[width=\textwidth]{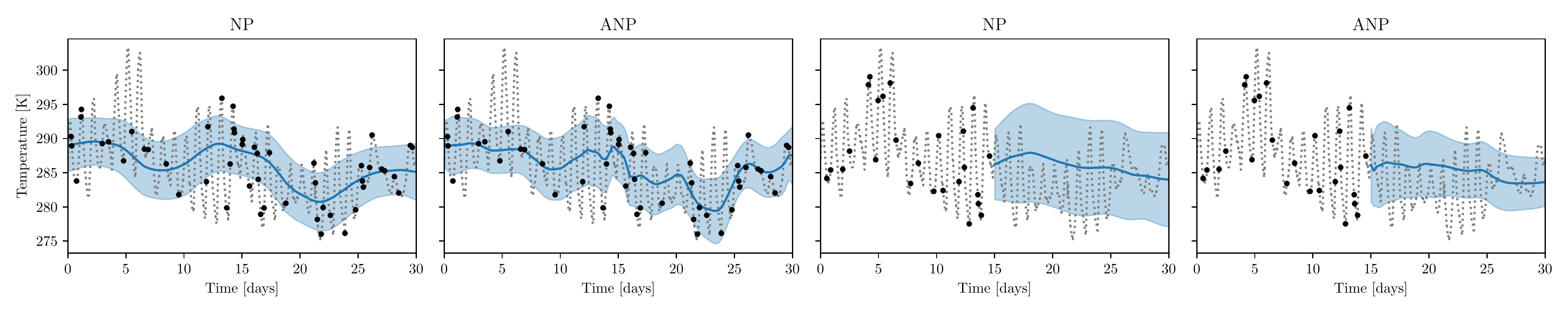}
    \caption{Examples from the temperature time series test set for NP and ANP. For the interpolation task (left) we provide context points from the full sequence, for the extrapolation task (right) we provide context points in the first half of the sequence and evaluate the second. Both methods are unable to fit the context points, likely because the frequency is too high to be represented in the models.}
    \label{fig:appendix:temperature_np}
\end{figure*}

\begin{figure*}
    \centering
    \includegraphics[width=0.95\textwidth]{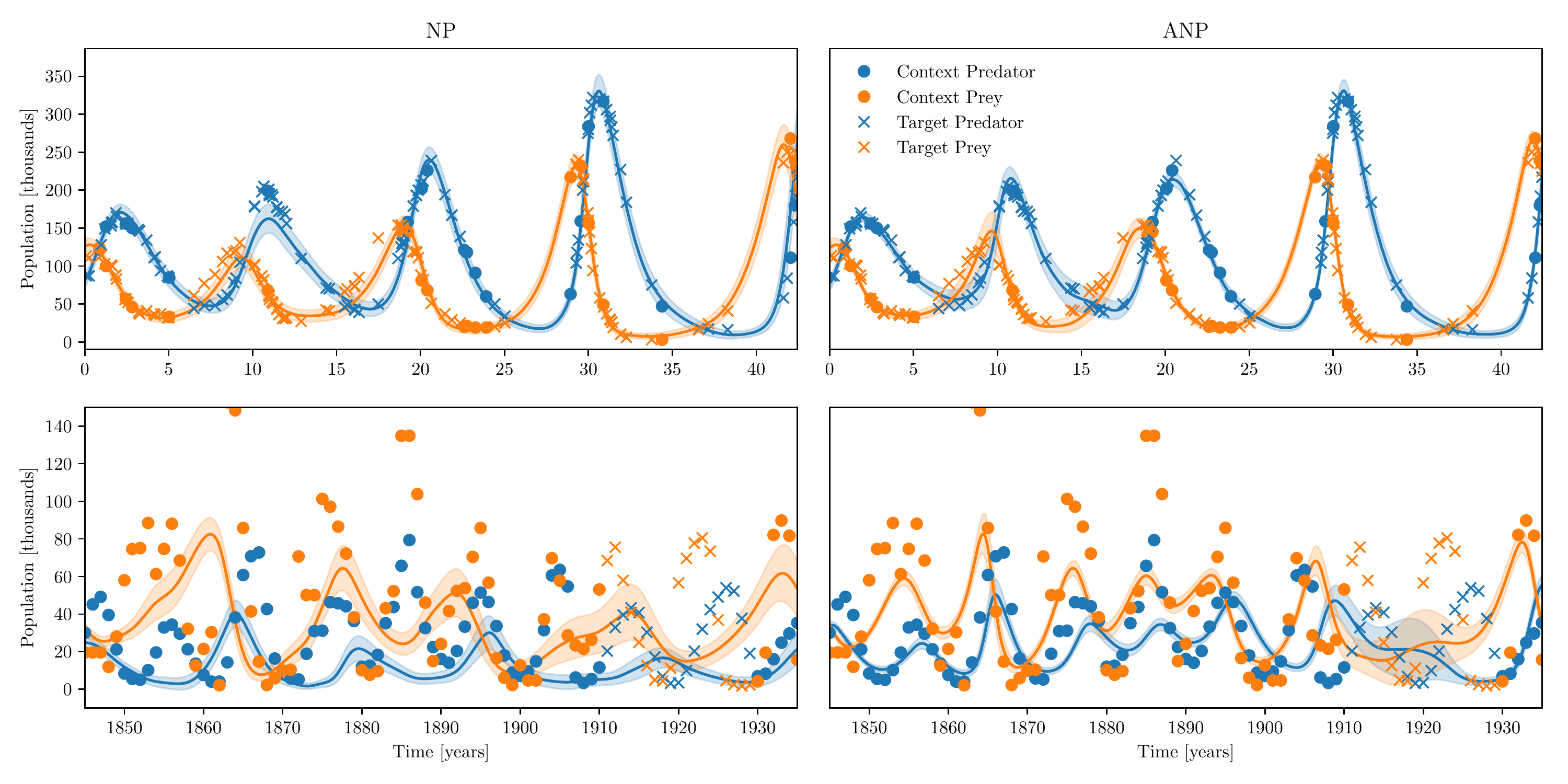}
    \caption{Example of NP and ANP applied to the simulated Lotka-Volterra population dynamics (top) and to the real Hudson Bay Company lynx-hare dataset (bottom). Similar to \textsc{ConvCNP} and \textsc{GP-ConvCNP}, seen in \cref{fig:population}, both work well on simulated data. On the real data, however, both struggle to fit the context points and produce a poor prediction for the test interval.}
    \label{fig:appendix:population_np}
\end{figure*}

In this section we show some additional results, specifically we show the performance as a function of the number of context points (\cref{fig:appendix:synthetic_numcontext}) as well as examples for NP and ANP on the temperature time series dataset in \cref{fig:appendix:temperature_np} and on the population dynamics dataset in \cref{fig:appendix:population_np}. We also show results on the population dynamics dataset in \cref{tab:appendix:population}, using a different evaluation method compared to the main manuscript.

\section{Image Experiments}
\label{sec:appendix:imageexperiments}

For a more complete comparison of our model with \textsc{ConvCNP}, we include image experiments, specifically MNIST, CIFAR10 and CelebA. For the latter two, we work with resampled images at $32^2$ resolution. The context set has a size drawn from $[20, 400)$ ($[20, 300)$ for MNIST), the target set a size drawn from $[50, 400)$, and we reconstruct both target and context points during training. We evaluate the average log-likehood of the model predictions on the respective test sets, as seen in \cref{tab:image}. The implementation of \textsc{ConvCNP} is again taken directly from the official repository, and we leave the architecture unchanged with the exception of swapping the kernel interpolation for a GP to make the comparison fair. All other hyper parameters are the same as in the time series experiments.

\begin{figure*}
    \centering
    \includegraphics[width=0.8\textwidth]{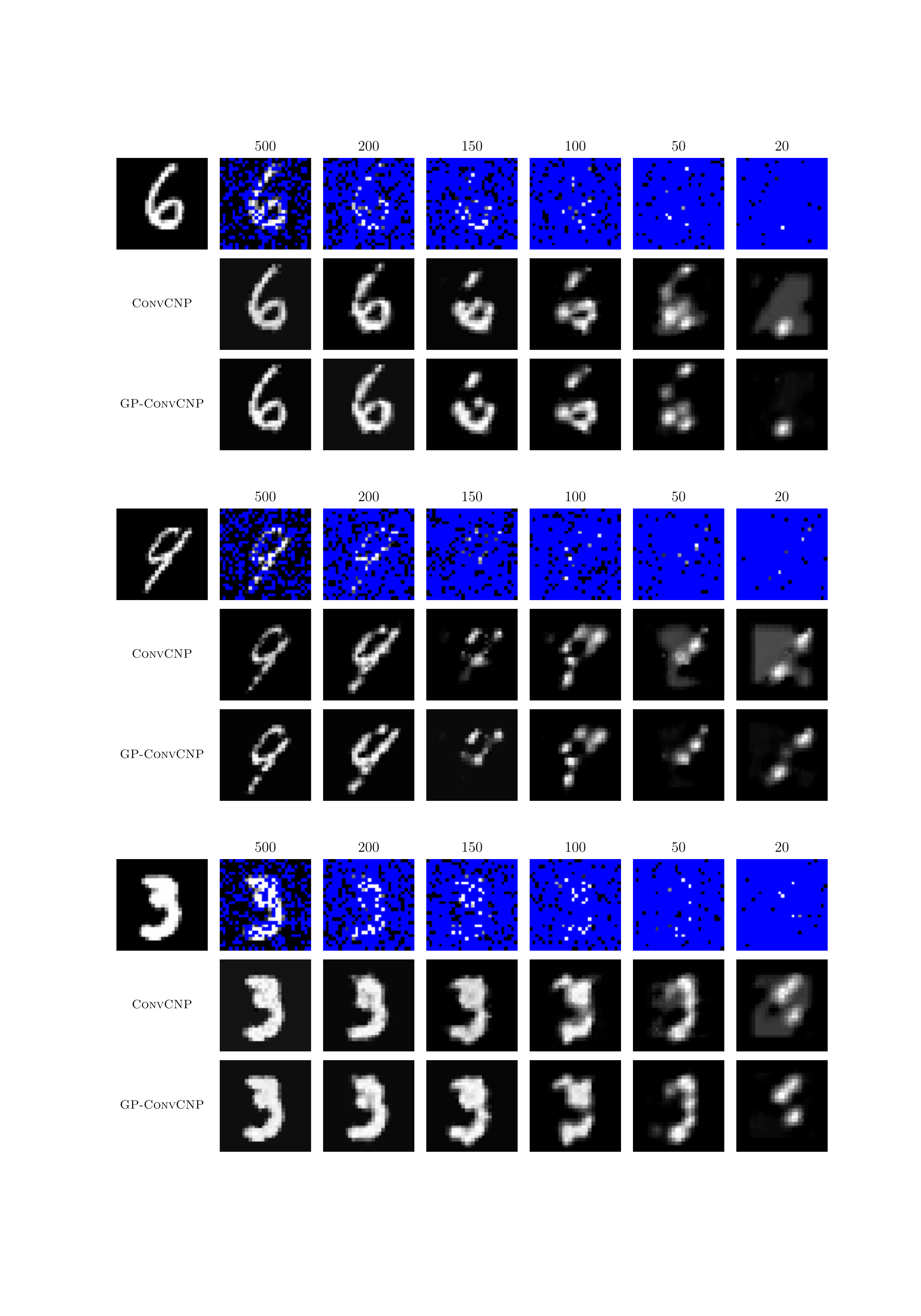}
    \caption{Examples for \textsc{ConvCNP} and \textsc{GP-ConvCNP} applied on MNIST test data. Models were trained on the training set. Numbers indicate the number of context points and the top left panel shows the reference image for each case.}
    \label{fig:appendix:mnist}
\end{figure*}

\begin{figure*}
    \centering
    \includegraphics[width=0.8\textwidth]{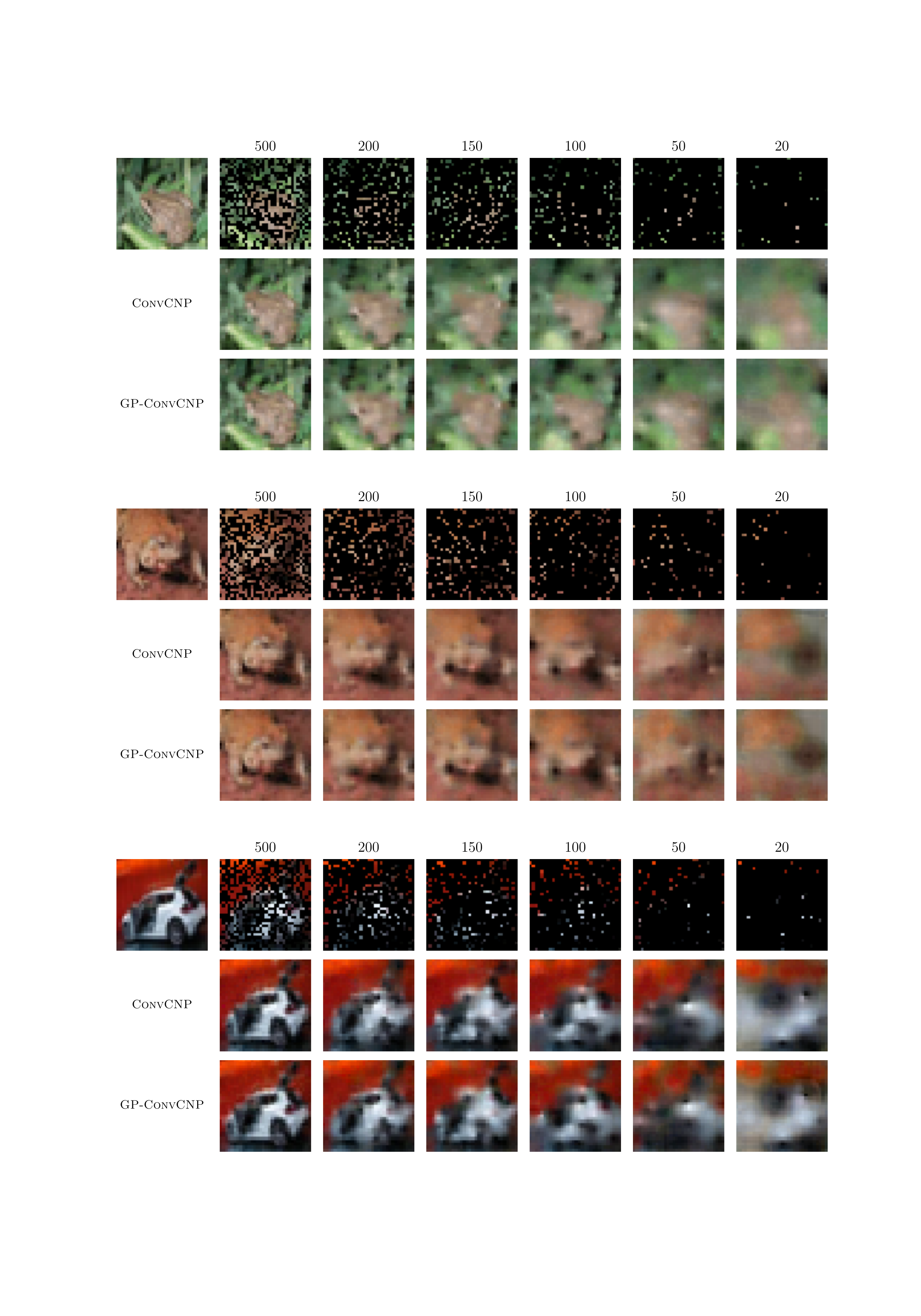}
    \caption{Examples for \textsc{ConvCNP} and \textsc{GP-ConvCNP} applied on CIFAR10 test data. Models were trained on the training set. Numbers indicate the number of context points and the top left panel shows the reference image for each case.}
    \label{fig:appendix:cifar10}
\end{figure*}

\begin{figure*}
    \centering
    \includegraphics[width=0.8\textwidth]{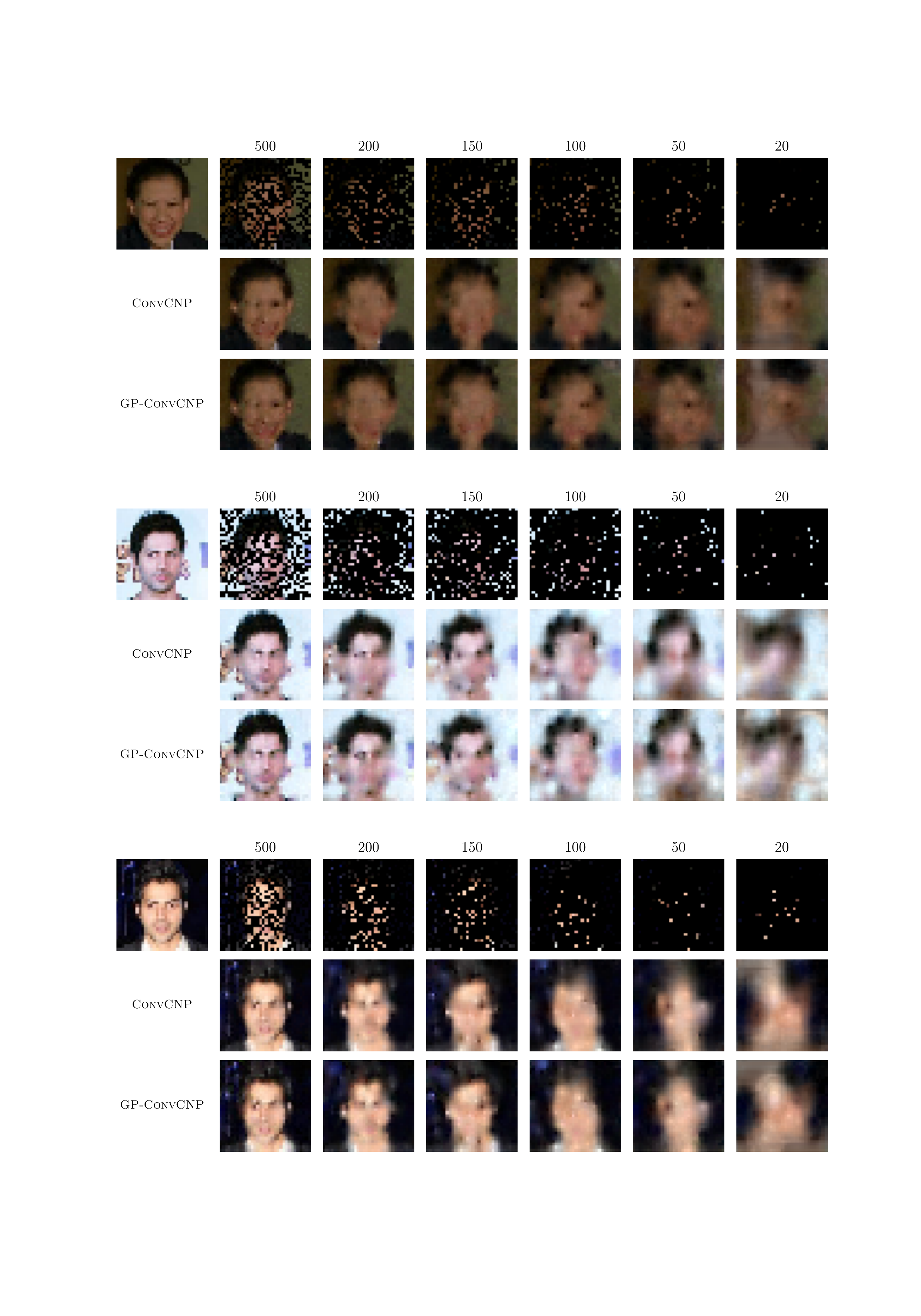}
    \caption{Examples for \textsc{ConvCNP} and \textsc{GP-ConvCNP} applied on CelebA test data, resized to 32x32 resolution. Models were trained on the training set. Numbers indicate the number of context points and the top left panel shows the reference image for each case.}
    \label{fig:appendix:celeba32}
\end{figure*}

Examples for both \textsc{ConvCNP} and \textsc{GP-ConvCNP} can be seen in \cref{fig:appendix:mnist}, \cref{fig:appendix:cifar10} and \cref{fig:appendix:celeba32}, with each example taken from the test sets. There is not noticeable visual difference between the two model, so we assume that the improved performance is due to better estimates of the predictive uncertainty (i.e. the standard deviation of the predicted Gaussian). In terms of performance, we found that inference takes roughly 1.5x as long for \textsc{GP-ConvCNP} as it does for \textsc{ConvCNP}, which we believe is still an acceptable tradeoff.

\end{document}